# Towards Democratized Flood Risk Management: An Advanced AI Assistant Enabled by GPT-4 for Enhanced Interpretability and Public Engagement

Rafaela Martelo[1], Kimia Ahmadiyehyazdi[2], Ruo-Qian Wang[3]
[1] Department of Mathematics, Rutgers University
[2] Institute for Water Resources Science & Technology, Texas A&M University, San Antonio
[3] Department of Civil and Environmental Engineering, Rutgers University

**Abstract**
Real-time flood forecasting is vital for effective emergency responses, but bridging the gap between complex numerical models and practical decision-making remains challenging. Decision-makers often rely on experts, while the public struggles to interpret flood risk information. To address this, we developed a customized AI Assistant powered by GPT-4. This tool enhances communication between decision-makers, the public, and forecasters, requiring no specialized knowledge. The framework leverages GPT-4's advanced natural language capabilities to search flood alerts, answer inquiries, and integrate real-time warnings with flood maps and social vulnerability data. It simplifies complex flood zone information into actionable advice. The prototype was evaluated on relevance, error resilience, and contextual understanding, with performance compared across different GPT models. This research advances flood risk management by making critical information more accessible and engaging, demonstrating the potential of AI tools like GPT-4 in addressing social and environmental challenges.

## 1. Introduction

As a critical component of flood risk management, effective communication of flood risks is important for at-risk populations to be prepared for emergency situations (Maskrey et al., 2016; Argent et al., 2016). A key tool in flood risk communication is flood maps, which are used to enhance public understanding by highlighting flood impacts and promoting preparedness (Arrighi et al., 2019; Henstra et al., 2019). However, traditional risk maps often fall short in decision support, as non-experts may not effectively utilize or understand them, leading to a gap in public awareness about flooding impacts (Leskens et al., 2014). Specifically, the challenges with traditional flood risk maps include 1) their failure to keep a balance between simplicity and complexity for public usability (Hagemeier-Klose and Wagner, 2009), 2) the difficulty in interpreting probabilistic risk information, such as flood return periods (O'Sullivan et al., 2012; Strathie et al., 2017), and 3) many people lack direct experience with floods, which hampers their ability to visualize and understand flood risks, especially in areas protected by dikes (Bradford et al., 2012; Botzen et al., 2009; Bubeck et al., 2012).

Furthermore, the traditional method of communicating flood risks through maps often follows a top-down strategy. This means that stakeholders are seen merely as receivers of information, with little to no involvement in creating the flood maps themselves. This approach can lead to flood maps that don't align with the actual needs of the users, since the content is not tailored to those it is meant to serve. This disconnect is a significant hurdle in communicating flood risks effectively (Meyer et al., 2012). A user-specific approach is required to improve the relevance and utility of flood maps. Furthermore, Rollason et al. (2018) highlight that stakeholders are particularly interested in understanding how predicted flood levels might affect their properties specifically, and how these predictions compare to previous flooding events. This underscores the need for flood communication to be more interactive and responsive to user needs.

Risk communication is also a key factor that shapes how people perceive and respond to flood risks (Hagemeier-Klose and Wagner, 2009; Strathie et al., 2017). The resilience of a community post-flood is closely tied to human perception, which is itself influenced by the social context surrounding the flood event (Bodoque et al., 2016, 2019), because they link emotions, risk perceptions, and behavioral responses to floods. Understanding how people—be it residents, businesses, or government bodies—perceive and comprehend flood risks is vital to the decisions they make and the actions they anticipate.

In recent years, flood risk management shifts towards incorporating real-time forecasting, which has added further complexity to the field practice. To effectively merge real-time forecasting with current practices, there's a need for more rigorous and prompt coordination of emergency responses, as well as more dynamic engagement with the public. Traditional decision-making processes are based on complex numerical models that, while accurate after calibration and validation, can introduce delays that impede rapid decision-making. The complexity is amplified by the dynamic mixture of socio-cultural, institutional, and environmental factors that create hurdles in the clear communication of flood risks and hamper public participation. Furthermore, demographic shifts and the intensifying effects of climate change aggravate these challenges, underscoring the urgent demand for innovative technology solutions. Such advancements are necessary to strengthen the resilience of communities to the growing menace of flood-related disasters.

The invention of Artificial Intelligence, particularly Large Language Models (LLMs), introduces a revolutionary opportunity with the potential to transform flood risk management. These models have the remarkable ability to generate text that mirrors human writing, producing everything from poetry (Köbis & Mossink, 2021) to political statements (Bullock & Luengo-Oroz, 2019) and scholarly papers (Hu, 2023). The application of Generative AI extends beyond text, as it has been used to create visual art (Gillotte, 2019), synthetic human faces (Whittaker et al., 2020), and even detailed magnetograms of the Sun (J. Liu et al., 2021), demonstrating its extensive range of capabilities across various disciplines. LLMs hold promise as a versatile tool that could bridge the gap between the complex integration of expert numerical modeling and the practical needs of non-expert stakeholders. This could enable more effective communication and decision-making processes in flood risk management by providing an accessible platform for public engagement and understanding. The application of LLM has the potential to enhance the experience of flood map users, provide an interactive platform for two-way communication, improve the efficiency and effectiveness of integrating real-time flood forecasting, and, more importantly, reshape the perception of flood risk and preparedness.

It is worth noting that AI has been employed in flood research for several years (Wang, 2021). Initially, the goal was to mine flood-related data from unstructured sources like social media. Early efforts (e.g. Wang et al., 2018) showcased the ability of computer vision combined with social media data to pinpoint flood locations. This approach benefits from the widespread adoption of smartphones and social media, transforming these platforms into effective tools for gathering real-time, ground-level data. Although concerns about the reliability of social media-based flood monitoring have been raised, particularly after Twitter (now X) ceased its precise location-sharing feature (Hu and Wang, 2020), AI has proven to be a valuable asset in improving flood modeling and mapping. In addition to data mining, AI has been used in predictive modeling (Chang et al., 2020), risk assessment (Pham et al., 2021), mitigation planning (Ye et al., 2021). These efforts focus on providing data to decision-makers, but studies engaging decision-makers in two-way communication in the modeling-decision cycle are still rare. This is due to the difficulty in the past when modeling systems could not adaptively respond to unstructured, random questions from decision-makers. The rise of LLM may provide a promising solution.

This paper presents our study to leverage the untapped capabilities of the LLM, i.e., Generative Pretrained Transformer (GPT-4) model, to create a specialized AI Assistant for flood risk assessment and proactive measures. GPT-4 is renowned for its state-of-the-art performance in various professional and academic benchmarks, often surpassing other LLM in a range of natural language processing tasks (OpenAI et al., 2023). In this study, we explore the untapped potential of GPT-4 to construct a specialized AI Assistant tailored for flood-related inquiries. As we delve into the unexplored realm of real-time flood forecasting and management, the implementation of GPT-4 marks a pioneering approach. While GPT-4's effectiveness across various sectors such as education, medical, history, mathematics, and physics is well-documented, its adaptation to the complex and diverse scenarios of flood risk management remains an area ripe for exploration (Liu et al., 2023).

We leverage the distinctive function calling feature of GPT-4 to equip our Flood AI Assistant with specialized capabilities of flood map inquiry, interpretation, and semantic search. The Flood AI Assistant adeptly integrates the function calling mechanism to access GIS APIs and query a specific social vulnerability dataset, utilizing GPT's ability to generate coordinates across different coordinate systems. In the following, Section 2 introduces our methodology, the research results are presented in Section 3, and a variety of applications are introduced in Section 4. We discuss the limitation and future research directions in Section 5.

## 2. Method

### 2.1 AI Assistant Framework

To develop such an AI Assistant, we designed four specialized functions, namely:

(1) *get_flood_map:* a function that provides valuable flood risk information by generating both a static flood map and an interactive map using standard latitude and longitude coordinates. The static map is color-coded with FEMA Flood Zone designations, enabling users to assess the risk associated with different regions. In case of any issues, an error message will be returned, ensuring transparency about potential retrieval errors for the static map. The interactive map allows users to explore flood risk dynamically.

(2) *get_flood_data:* a function that returns information related to flood risk and property details based on provided addresses. The response includes details such as flood hazard area, FEMA FIRM (Flood Insurance Rate Map) panel information, elevation, political and community information, census data, and geocoding details like latitude and longitude.

(3) *get_svi_stats_and_tracts:* a function retrieves Social Vulnerability Index (SVI) statistics and information about census tracts based on user-specified conditions, including state abbreviation, county, SVI theme, a comparison operator, and a threshold value. It answers questions related to the vulnerability of specific areas by providing statistics such as the total number of vulnerable areas, maximum SVI score, minimum SVI score, and average SVI score, along with details (including boundary coordinates) for each relevant census tract in JSON format. With the interface, the AI Assistant can also display an interactive map with the census tracts of interest for the user to explore. In the prototype we're using the 2020 US SVI data (*CDC/ATSDR SVI data and documentation download*, 2022).

(4) *get_flash_flood_warnings:* a function retrieves active flood alerts from the National Weather Service's API based on the specified location. It returns details about the flood alerts, including the event type, description, timestamp sent, timestamp expires, and area description.

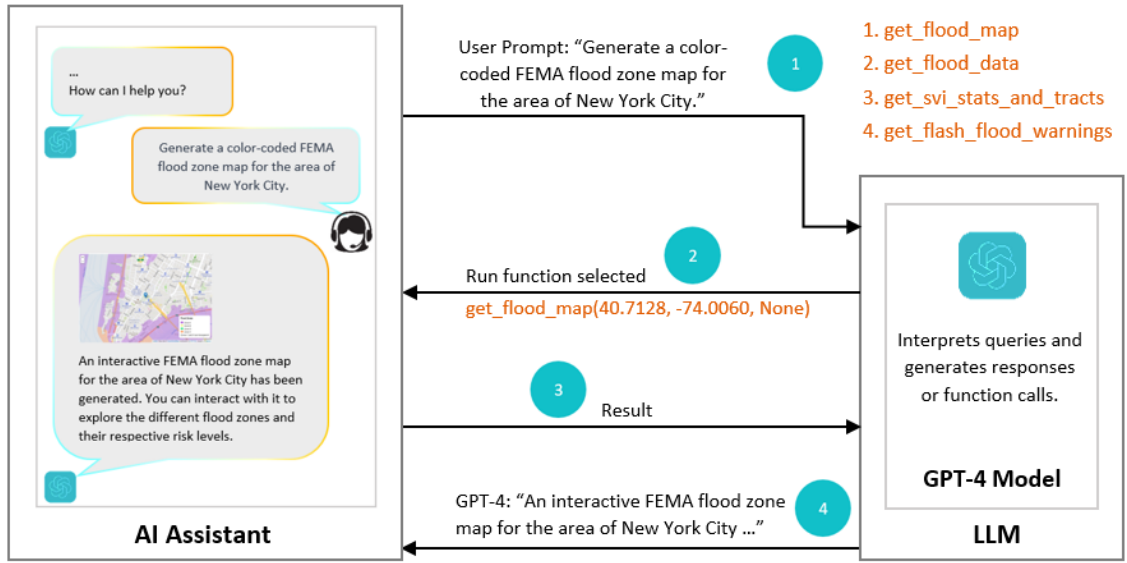

**Figure 1:** Diagram Illustrating GPT-4 API Interaction with Function Calling in the Flood AI Assistant Prototype

With these functions, we design the framework of the AI Assistant This diagram (Figure 1) provides an overview of how the GPT-4 API processes user prompts with the aid of specialized function calls within the Flood AI Assistant:

1) **User Prompt Submission:** The chat application sends user's queries, along with a list of available functions, to the GPT-4 API.

2) **Model Response and Decision:** The GPT-4 model evaluates the query and either generates a text response directly or identifies the need for a specific function call.

3) **Function Execution:** If a function call is required, the AI Assistant executes the requested function, harnessing the relevant data or processing the necessary computation, and then returns the results to the GPT-4 API.

4) **Response Generation and Iteration:** Utilizing the data from the executed function, the model crafts a coherent text response. In certain cases, the API may initiate a subsequent function call for additional information or clarification.

## 2.2 Evaluation Methodology

Our study evaluated the performance of our flood-related AI prototype using a structured approach. The evaluation focused on six key criteria, categorized under three primary factors: Relevance, Error Resilience, and Context Understanding.

**Performance Criteria:**

- **Relevance**
  - *"Accuracy"* assesses if the AI's responses are factually correct, relevant to the query, and free from misinformation.
  - *"Completeness"* determines whether the response fully addresses all aspects of the query.
- **Error Resilience**
  - *"Error Handling"* evaluates the AI's capability to manage errors, uncertainties, or incomplete data, including its ability to recognize and communicate these issues.
  - *"Informative Responses"* rates the AI's effectiveness in providing useful information, especially when data is unavailable, or the query is ambiguous.
- **Context Understanding**
  - *"Appropriateness"* checks if the response is suitable and relevant to the query's context.
  - *"Adaptability"* measures the AI's flexibility in modifying responses based on the specifics of the query or evolving conversation.

These criteria were applied to assess the prototype's responses to a set of predefined prompts.

**Prompt Categories:**

1. Get_Flood_Map Function
2. Get_Flood_Data Function
3. Get_SVI_Stats_And_Tracts Function
4. Get_Flash_Flood_Warnings Function
5. General Understanding of Flood-Related Queries

Each category was designed to test the prototype's specialty functions and its general comprehension of flood-related scenarios, leveraging the GPT-4 model's pre-trained knowledge base.

We conducted two iterations of testing with six prompts in each of the five prompt categories. The original prompts were designed to test the performance of both ambiguous and more specific queries for each prompt category. The initial iteration was used to identify prompts requiring refinement to maximize the prototype's use of its specialty functions. Iteration 2 was used to adjust the initial prompt and evaluate the performance. Each response was scored on a scale of 1 to 5 for all criteria, with the following qualitative interpretations:

- 1/5 (Poor): Significantly unsatisfactory in meeting the criteria. The response may be largely irrelevant, inaccurate, or not useful.

- 2/5 (Fair): Partially meets the criteria but has notable deficiencies. The response might be somewhat relevant or useful but lacks key elements or clarity.
- 3/5 (Good): Adequately meets the criteria. The response is relevant and useful, but there is room for improvement in certain aspects.
- 4/5 (Very Good): Strongly meets the criteria with only minor shortcomings. The response is highly relevant, accurate, and largely fulfills the user's needs.
- 5/5 (Excellent): Exceptionally meets the criteria. The response is comprehensive, highly accurate, thoroughly informative, and fully addresses the user's query.

**2.3 Computational configuration**

For the evaluation, we utilized the "gpt-4-1106-preview" model, which operates under the limitation of returning a maximum of 4,096 output tokens and uses default temperature settings. This limitation on token output is crucial as it necessitates the management of conversation length. When the conversation exceeds this token limit, the model automatically omits the last message to fit within these constraints, leading to a potential loss of context. It's also important to note that the model's training data includes information up to April 2023, which defines the extent of its knowledge base. The default temperature setting influences the AI's response generation, balancing creativity and relevance to the query. Understanding these limitations and settings is essential for interpreting the AI's responses accurately and evaluating its performance in addressing flood-related queries and effectively utilizing its specialty functions.

This approach in evaluation was designed to not only test the effectiveness of the AI model in responding to flood-related queries but also to explore the practical applications and limitations of its specialty functions. Understanding the interface and structure of function design is critical for optimizing the AI Assistant's performance. This evaluation helped uncover the strengths and weaknesses of the model, providing valuable insights into how its functionalities can be enhanced for efficiency and effectiveness. The insights gained from this evaluation are intended to guide future developments, ensuring that subsequent iterations of the AI Assistant are more adept at handling flood-related inquiries and providing accurate, relevant, and comprehensive information. This will enable the development of a more efficient and useful tool for flood risk management and emergency response planning.

The evaluation of the model involved designing *one-shot* prompts to test the specialty functions under specific, ambiguous, or vague queries. The numbering system identifies the prompt category, while single, double, or triple letters indicate the stages of prompt refinement. For example, "3B" is the second iteration of prompt B in Category 3, and "3BB" is a further refinement of "3B." Detailed information on these prompts, their responses, and individual criteria scores are available in Appendix A, Tables A1 and A2. It is worth noting our analysis focuses on Iteration 2, because Iteration 1 is preliminarily experimental for prompt refinement and consistency examination.

With original prompts, Categories 2, 4, and 5 did not require prompt refining to thoroughly study the prototype's performance with these categories. Category 2 demonstrated the model's competence in responding to queries for flood data for specific addresses and its ability to request additional information. Category 4, focusing on flood alerts, showcased the model's performance with original prompts, so further refinement is unnecessary. For general flood-related queries, i.e., Category 5, the model exhibited a robust understanding and handling of a diverse range of topics.

2.4 Comparison with other GPT models

The latest GPT models were used to assess the accuracy and the performance of the output for a comparative study against the main model used in the study (GPT-4-1106-preview). The response evaluation is the same as using the metrics above for two GPT models: GPT-4o ("o" for "omni") is the most advanced multimodal model, that's faster and cheaper than GPT-4 Turbo with stronger vision capabilities — it generates text two times faster and is 50% lower in token costs. GPT-4o-mini a cost-efficient small model, which is more intelligent and cost-efficient than GPT-3.5 Turbo with vision capabilities. GPT-4o-mini was recommended to replace GPT-3.5-turbo.

3. Illustrative Examples

Here we present two illustrative examples to showcase the inputs and outputs for the system. These cases are aimed to give a concrete and specific overview of the system application to facilitate the understanding of the evaluation results.

Case Study 1 was designed to demonstrate the top-bottom scenario of flood information inquiry. Specifically, the questions started with an inquiry of the general region and then asked for information about specific addresses. The AI Assistant adeptly navigates the questions, transitioning seamlessly into historical flood impact analysis and culminating with detailed Social Vulnerability Index statistics for specific neighborhoods. Case Study 2 was designed to test the bottom-top scenario. The questions start from a specific address and transit to the general area and specific mitigation actions. The Assistant was shown able to evaluate flood risks for a construction site in Baton Rouge. It identifies the flood zone, provides a flood map with comprehensive mitigation strategies for the site.

These scenarios exemplify the AI Assistant's remarkable capacity to synthesize diverse and unstructured flood-related data into coherent, actionable insights. Whether it's merging real-time flood alerts with historical data and social vulnerability indices or translating flood zone categorizations into practical construction advice, the Assistant proves invaluable. Its ability to guide users from problem identification through to solution formulation is a testament to its potential as a transformative tool in flood risk management. By offering a nuanced understanding that combines technical data with socio-economic considerations, the AI Assistant empowers users to make well-informed, context-sensitive decisions, paving the way for more resilient and equitable responses to flood risks.

**Case Study 1:** New Orleans Flood Alert and Vulnerability Assessment.

What are the current flood alerts for the New Orleans area?

Currently, there are no flash flood warnings issued for the New Orleans area in Louisiana. Please note that flood situations can change rapidly, so it's always a good idea to stay informed by checking for updates regularly, especially during heavy rain events or severe weather conditions. If you need further assistance or more detailed information, feel free to ask.

Can you identify the neighborhoods in New Orleans that have historically faced the most severe flooding?

To identify the neighborhoods in New Orleans that have historically faced severe flooding, we can consider areas within Special Flood Hazard Areas (SFHAs), as well as those affected by hurricanes and severe storms in the past. Flood maps, historical data, and local knowledge can provide us with this information.

I can provide you with a general idea of the neighborhoods based on historical events, notably from Hurricane Katrina, which hit New Orleans in 2005. Some of the neighborhoods that were most severely affected by flooding at that time include:

- The Lower Ninth Ward: This neighborhood suffered extensive damage due to the levee failures, with many homes completely destroyed by the floodwaters.

- Lakeview: Located near Lake Pontchartrain, Lakeview also experienced severe flooding from levee breaches.

- New Orleans East: This area was inundated due to its proximity to the Industrial Canal and the Gulf Intracoastal Waterway.

- Gentilly: This neighborhood, located on the edge of the city, is low-lying and was significantly impacted by flooding.

- St. Bernard Parish: Although not a neighborhood of New Orleans, St. Bernard Parish, adjacent to the city, was almost entirely flooded.

Current flood maps can also show areas that are at risk of flooding based on latest data. I can fetch a flood map for New Orleans to show the designated flood zones, which will give us a visual representation of the areas at risk. Would you like to proceed with this?

For the neighborhoods you mentioned, can you provide SVI statistics for areas with an SVI score higher than 0.9?

**Case Study 2:** Baton Rouge Construction Site Flood Risk Analysis.

Is there a flood risk for the proposed construction site at 777 Ben Hur Rd Baton Rouge, LA 70820 in Baton Rouge?

The construction site at 777 Ben Hur Rd, Baton Rouge, LA 70820 is located within a flood zone designated as "AE." This classification indicates that the area is at a high risk for flooding, as it is within the Special Flood Hazard Area (SFHA), where there is a 1% annual chance of flood hazard. The designation "T" for SFHA_TF also affirms that it is in a regulatory floodplain.

These zones are areas subject to inundation by the base flood, which is the flood having a 1% chance of being equaled or exceeded in any given year. It's also known as the 100-year flood. Construction in this area will likely need to comply with local and federal floodplain management regulations, such as raising buildings above the base flood elevation.

It's important for anyone planning construction in this area to consult with local building officials, FEMA floodplain maps, and potentially an engineer specializing in hydraulics or hydrology for specific construction requirements and mitigation strategies.

Can you provide a detailed flood zone map for this area to help us understand the specific flood risks?

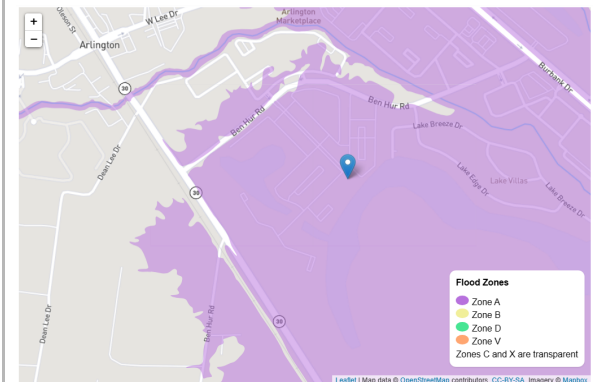

I apologize for the inconvenience; there seems to be an internal error while retrieving the static flood map. However, you can view the interactive flood map for the location at 777 Ben Hur Rd, Baton Rouge, LA 70820 by following the link provided below. This map will show the flood zones and help you understand the specific flood risks for the area.

[View Interactive Flood Map for 777 Ben Hur Rd, Baton Rouge, LA 70820](sandbox:/static/interactive_flood_map.html)

To get a detailed understanding of the flood risks in this area, you can also refer to FEMA's official flood maps, or contact local authorities for the most accurate and up-to-date information on flood zones and requirements for building in flood-prone areas.

Based on the flood risk and zone information, what mitigation measures should we consider for this construction project?

Given that the proposed construction site at 777 Ben Hur Rd, Baton Rouge, LA 70820 is within a flood zone designated as "AE," it is important to consider the following mitigation measures for the

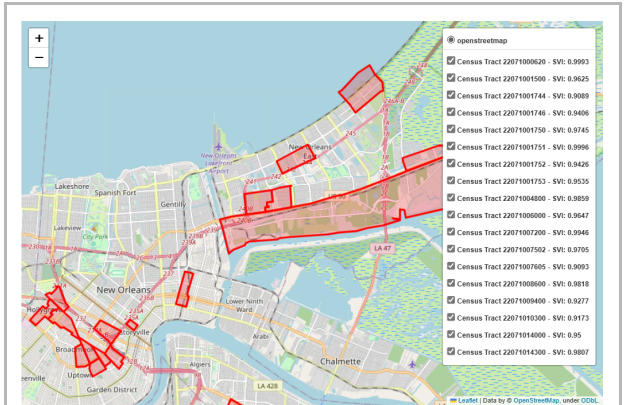

In the Orleans County area of New Orleans, there are several census tracts with SVI scores higher than 0.9, indicating a high level of social vulnerability. Here are the areas along with their respective SVI scores:

1. Census Tract 22071000620 - SVI Score: 0.9993
2. Census Tract 22071001500 - SVI Score: 0.9625
3. Census Tract 22071001744 - SVI Score: 0.9089
4. Census Tract 22071001746 - SVI Score: 0.9406
5. Census Tract 22071001750 - SVI Score: 0.9745
6. Census Tract 22071001751 - SVI Score: 0.9996
7. Census Tract 22071001752 - SVI Score: 0.9426
8. Census Tract 22071001753 - SVI Score: 0.9535
9. Census Tract 22071004800 - SVI Score: 0.9859
10. Census Tract 22071006000 - SVI Score: 0.9647
11. Census Tract 22071007200 - SVI Score: 0.9946
12. Census Tract 22071007502 - SVI Score: 0.9705
13. Census Tract 22071007605 - SVI Score: 0.9093
14. Census Tract 22071008600 - SVI Score: 0.9818
15. Census Tract 22071009400 - SVI Score: 0.9277
16. Census Tract 22071010300 - SVI Score: 0.9173
17. Census Tract 22071014000 - SVI Score: 0.95
18. Census Tract 22071014300 - SVI Score: 0.9807

These tracts represent areas within New Orleans that are more likely to need support during emergencies, including flooding, due to factors such as socioeconomic status, household composition, race, ethnicity, and housing/transportation characteristics.

Please find the map below showing the highlighted census tracts with high SVI scores:

[Interactive SVI Map with Census Tracts Highlighted](sandbox:/static/svi_tracts_and_stats_map.html)

Keep in mind that the SVI scores reflect more than just flood risk—they encompass various aspects of vulnerability that can heighten the impact of disasters.

construction project:

1. **Elevation**: Elevate the building above the base flood elevation (BFE) as outlined by FEMA for zone AE. This might involve the use of stilts, piers, or fill to raise the structure's lowest floor.

2. **Foundation**: Design a foundation that can resist flood forces, such as deep foundations that extend to stable soils below the zone of influence of floodwaters, or pile foundations that can resist lateral forces.

3. **Floodproofing**: Make the building watertight below the BFE through methods like sealing walls with waterproof coatings, using flood-resistant materials, installing backflow valves to prevent sewage system backups, and using sump pumps.

4. **Site Planning**: Orient and position the building to minimize exposure to flowing water. Preserve natural features that absorb floodwaters, and avoid construction in areas that would be cut off by flooded access routes.

5. **Stormwater Management**: Implement green infrastructure, like rain gardens and permeable pavement, to manage runoff and reduce the potential for flooding on and around the site.

6. **Flood Barriers**: Consider temporary flood barriers or levees that can be deployed when flooding is imminent to protect the structure and site.

7. **Utilities and Services**: Elevate and secure mechanical, electrical, and plumbing systems above the BFE, and use flood-resistant materials for components that cannot be elevated.

8. **Emergency Action Plan**: Develop an emergency response plan for occupants that outlines evacuation routes, procedures, and safety measures in case of a flood.

9. **Landscaping**: Utilize landscaping to help divert floodwaters away from buildings and reduce erosion.

10. **Insurance**: Ensure that the building is adequately insured under the National Flood Insurance Program (NFIP), especially since buildings within the AE zone are usually required to have flood insurance if they are financed by a federally insured or regulated lender.

11. **Regulatory Compliance**: Comply with local floodplain management ordinances and zoning codes, which may have stricter requirements than the federal regulations.

Incorporating these measures can help protect the building and its future occupants from flood risks. Consulting with a civil engineer specializing in hydraulics and flood prevention, as well as local authorities, will be crucial for developing an effective flood mitigation strategy that meets all local, state, and federal requirements.

**Figure 2:** Example conversation records of the case studies

## 4. Evaluation Results

### 4.1 Performance Evaluation

In this section, we computed the average scores across criteria per prompt categories to evaluate the prototype's performance in each predefined function and general flood-related knowledge (Table 1). For Categories 1 and 3, both the original and refined prompt scores were included to reflect the shortcomings of these functionalities. Error Resilience emerged as the lowest-scoring criteria across the predefined functions, primarily due to inadequate informative responses in error situations or when handling vague queries. This factor also influenced the adaptability of the Assistant, impacting its completeness score. Notably, the categories of general flood understanding, and flash flood alerts exhibited the highest performance, achieving overall scores of 5 and 4.3, respectively. In contrast, the flood map and SVI stats functions demonstrated lower performance, both achieving an overall score of 4. These findings suggest that enhancing function design for clarity, ease of use, and efficient token usage will be crucial in future development.

| Prompt Category | Relevance | | Error Resilience | | Context Understanding | | Overall Performance |
|---|---|---|---|---|---|---|---|
| | Accuracy | Completeness | Error Handling | Informative Responses | Appropriateness | Adaptability | |
| 1 | 2.33 | 1.66 | 1.83 | 1.83 | 2 | 1.83 | 1.91 |
| | 1.99 | | 1.83 | | 1.91 | | |
| 2 | 4.5 | 3.66 | 4.16 | 3.66 | 4 | 3.5 | 3.91 |
| | 4.08 | | 3.91 | | 3.75 | | |
| 3 | 4.83 | 4.33 | 3.33 | 4.83 | 3.83 | 4.33 | 4.24 |
| | 4.58 | | 4.08 | | 4.08 | | |
| 4 | 4.5 | 4.16 | 4.16 | 4.5 | 4 | 3.83 | 4.19 |
| | 4.33 | | 4.33 | | 3.91 | | |
| 5 | 5 | 5 | 4 | 5 | 5 | 4 | 4.66 |
| | 5 | | 4.5 | | 4.5 | | |

**Table 1**: Average scores across all six criteria and mean overall performance grouped by prompts category.

## 4.2 Response Time Analysis

Detailed response times for each prompt, as observed during iteration 2, can be referenced in the "Observations" column of Table A2 in Appendix A. The information about the time consumption in each inquiry provides another dimension to view of the response dynamics across different categories and reflect the computational resource demands.

An insightful pattern was revealed in the AI system's performance. When a function call was involved, the average response time was recorded at 36 seconds. The prompt category for flash flood alerts (Category 4) demonstrated the fastest response within this group, taking only 5 seconds. In contrast, the longest response time was observed in the SVI stats category (Category 3), which took up to 198 seconds by average, especially in instances where the model attempted but failed to retrieve data. Conversely, prompts that did not invoke a function had a significantly lower average response time of 12 seconds. The quickest responses in this group, typically around 3 seconds, were often observed when the model requested additional details from the user. The slowest response in this category, at 37 seconds, was noted in the general flood-related knowledge category (Category 5), where the responses were generally more comprehensive and detailed. A deeper dive into the category-specific response times reveals that Category 3 (SVI stats) had the longest average response time for function calls, averaging 85 seconds, with a range from 57 seconds to an extreme case of 198 seconds. This discrepancy highlights that functions retrieving or searching large datasets can require more processing time, underscoring the need for efficiency in function design.

In terms of the overall speed of the AI system, the average response times can be considered satisfactorily fast, especially given the complexity of the tasks performed. The faster responses for simpler queries or when additional information is requested indicate the model's efficiency in handling straightforward tasks. However, the longer response times for more complex data retrieval tasks, like in the SVI stats category, suggest areas for optimization. Such response times, while still within reasonable bounds, could be improved with more efficient data handling or system enhancements.

It is important to contextualize these findings by acknowledging that response times can be influenced by external factors, such as server response time, internet speed, and GPT versions, which could account for some of the variability observed. These findings not only contribute to our understanding of the AI system's operational efficiency but also provide valuable insights for future improvements, particularly in optimizing the system for more rapid data processing without compromising the accuracy or comprehensiveness of the responses.

### 4.3 Comparison among GPT models

The average performance score and response time were used to compare the models in Figure 3a. GPT-4o showed the highest accuracy with a moderate response time. In contrast, GPT-4-1106-preview had the lowest accuracy and the slowest response time, while GPT-4o-mini was the fastest but with reduced accuracy, as expected. Figure 3b provides a more detailed comparison of response times, confirming that GPT-4o-mini consistently performed the fastest. These results suggest that GPT-4o is ideal when accuracy is the priority, while GPT-4o-mini excels in scenarios where quick responses are more important. On the other hand, GPT-4-1106-preview, as a preliminary model, requires further tuning to match the performance of the other models.

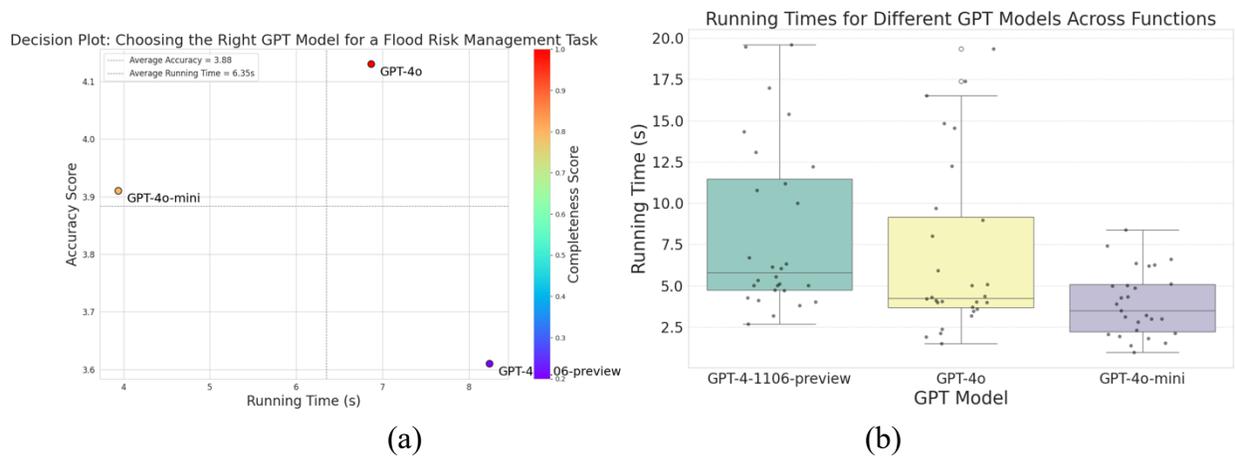

(a)                              (b)

**Figure 3:** Performance comparison among different models of GPT

## 5. Discussion

Our research, utilizing GPT-4 models for flood risk management, has unveiled the significant potential of AI Assistants in bridging the gap between intricate flood data and practical, accessible information. Despite the challenges posed by the model's token limitations and the inherent complexity of flood-related queries, our prototype AI Assistant demonstrated considerable proficiency in interpreting and responding to a diverse range of flood-related questions. This is evident in the successful handling of specific and ambiguous queries, as well as its ability to relay complex data in an understandable format.

The detailed evaluation of the AI Assistant, although conducted with limited trials, revealed key strengths and areas for improvement. Notably, the Assistant's performance in general flood knowledge queries shed lights on its use of predefined functions, underscoring the need for clearer function instructions and design enhancements. The study also highlighted the Assistant's struggle with large data queries and its limited adaptability in response to errors and vague queries. These insights are invaluable in guiding future iterations of the system.

We would also like to comment on the error handling capability of the AI assistant. In Category 1 (Get_Flood_Map Function), the function's shortage to retrieve static maps due to server issues provided a valuable opportunity to evaluate the model's error handling. We observed that while the model effectively communicated errors, it often missed opportunities to enhance user understanding, such as elaborating on color-coded flood map areas. Figure 4 shows mainly improved error resilience through prompt refinements, such as directly requesting specific locations or rephrasing prompts. However, Relevance and Context Understanding were also generally improved. The main improvement was seen in prompt D's refinement, while E only improved Error Resilience. Prompt F slightly improved across all three main criteria.

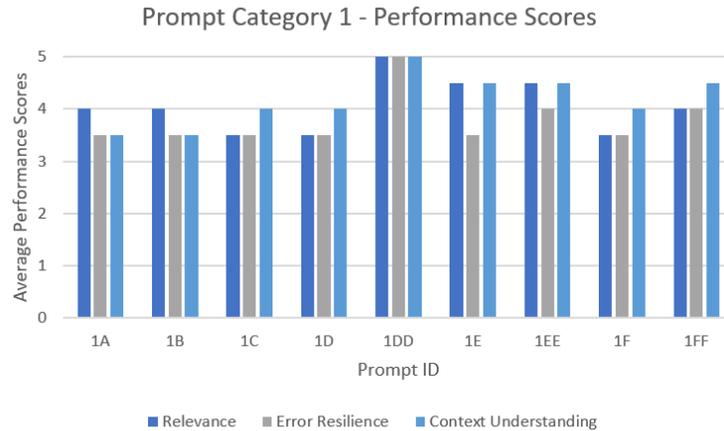

**Figure 4:** This graph illustrates the average performance scores across three main evaluation criteria: Relevance, Error Resilience, and Context Understanding. It specifically compares the performance improvement for prompts D, E, and F following their refinement. The data highlights how targeted adjustments to the prompts led to enhanced overall performance of the AI system, demonstrating its adaptability and responsiveness to more refined user inputs.

For Category 3 (Get_SVI_Stats_And_Tracts Function), the model faced challenges handling large data queries, often leading to the overlooks of crucial context or function response due to exceeding the maximum token output limit. To address this, we refined prompts to narrow down requested areas and social vulnerability scores. However, significant challenges remained due to the function's tendency to retrieve extensive data, which limits its practical utility. In real-world applications, alternative display methods for census tract data could be implemented to circumvent token limitations. Figure 5 illustrates that refining prompts to specify smaller score ranges or directly ask for a model-suggested area of interest notably improved the model's performance. The first refinement of Prompt D initially performed worse, but a second refinement considerably improved the performance, outperforming the original prompt. Refinements of prompts B, E, and F resulted in perfect scores in all 3 criteria. These observations highlight the importance of clear and informative function instructions for the model, and the need for better-designed functions overall.

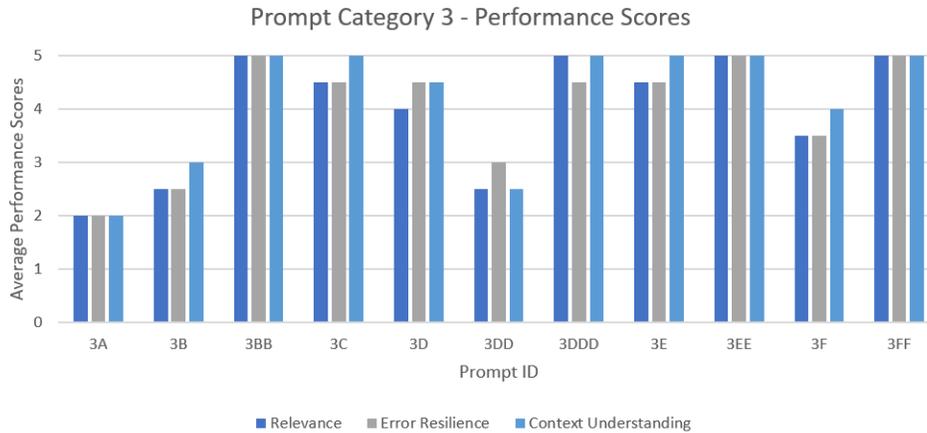

**Figure 5**: This graph illustrates the average performance scores across three main evaluation criteria: Relevance, Error Resilience, and Context Understanding. It specifically compares the performance improvement for prompts B, D, E, and F following their refinement. The data highlights how targeted adjustments to the prompts led to enhanced overall performance of the AI system, demonstrating its adaptability and responsiveness to more refined user inputs.

## 6. Conclusion and Future Work

Our investigation into AI-assisted flood risk management has unveiled promising insights, despite facing limitations and complexities inherent in flood-related inquiries. The prototype AI Assistant showcased a commendable ability to interpret and respond to a broad array of flood-related questions, ranging from highly specific to more ambiguous ones. It stood out in its capacity to transform complex data into understandable and actionable information, an essential attribute for informed decision-making in managing flood risks. However, the study highlighted challenges such as the model's struggles with handling extensive data sets and its limited effectiveness in navigating errors and vague queries. These observations highlight the potential and areas for improvement in AI tools for environmental risk management.

In future endeavors, our efforts will be geared towards enhancing the shortcomings while building on the identified strengths. Specifically, we will focus on enhancing the AI Assistant's functionality in handling extensive datasets and its resilience to errors and adaptability. We aim to integrate the Assistant with state-of-the-art tools for data retrieval and analysis, thereby augmenting its efficacy in real-time flood risk assessment. Our pursuit will involve not only refining current features but also exploring new avenues, such as GPT-4's visual input capabilities. This will allow the Assistant to process and interpret visual data, including maps, satellite imagery, and on-site photographs, to offer a more comprehensive understanding of flood scenarios. The overarching objective is to evolve the AI Assistant into a versatile and user-friendly tool that can cater to diverse needs, from aiding decision-makers in strategic planning to informing the general public about flood risks.

**Acknowledgment**

This study is funded by the Rutgers's Research Incubator in Climate and Health, Seed Funding Initiative and Research Council Award – "Engaged Climate Action". The source code and data are available through the GitHub repository: https://github.com/rqwang/FloodGPT_Prototype.

**References**

Agency for Toxic Substances and Disease Registry. (2022, October 26). *CDC/ATSDR SVI data and documentation download*. Centers for Disease Control and Prevention. https://www.atsdr.cdc.gov/placeandhealth/svi/data_documentation_download.html

Argent, R. M., Sojda, R. S., Giupponi, C., McIntosh, B., Voinov, A. A., & Maier, H. R. (2016). Best practices for conceptual modelling in environmental planning and management. *Environmental Modelling & Software*, *80*, 113–121. https://doi.org/10.1016/j.envsoft.2016.02.023

Arrighi, C., Pregnolato, M., Dawson, R. J., & Castelli, F. (2019). Preparedness against mobility disruption by floods. *Science of The Total Environment*, *654*, 1010–1022. https://doi.org/10.1016/j.scitotenv.2018.11.191

Bodoque, J. M., Amérigo, M., Díez-Herrero, A., García, J. A., Cortés, B., Ballesteros-Cánovas, J. A., & Olcina, J. (2016). Improvement of resilience of urban areas by integrating social perception in flash-flood risk management. *Journal of Hydrology*, *541*, 665–676. https://doi.org/10.1016/j.jhydrol.2016.02.005

Bodoque, J. M., Díez-Herrero, A., Amerigo, M., García, J. A., & Olcina, J. (2019). Enhancing flash flood risk perception and awareness of mitigation actions through Risk Communication: A pre-post survey design. *Journal of Hydrology*, *568*, 769–779. https://doi.org/10.1016/j.jhydrol.2018.11.007

Botzen, W. J., Aerts, J. C., & van den Bergh, J. C. (2009). Dependence of flood risk perceptions on socioeconomic and objective risk factors. *Water Resources Research, 45(10). https://doi.org/10.1029/2009wr007743*

Bradford, R. A., O'Sullivan, J. J., van der Craats, I. M., Krywkow, J., Rotko, P., Aaltonen, J., Bonaiuto, M., De Dominicis, S., Waylen, K., & Schelfaut, K. (2012). Risk perception – issues for flood management in Europe. *Natural Hazards and Earth System Sciences*, *12*(7), 2299–2309. https://doi.org/10.5194/nhess-12-2299-2012

Bubeck, P., Botzen, W. J., & Aerts, J. C. (2012). A review of risk perceptions and other factors that influence flood mitigation behavior. *Risk Analysis: An International Journal, 32*(9), 1481–1495. https://doi.org/10.1111/j.1539-6924.2011.01783.x

Bullock, J., & Luengo-Oroz, M. (2019). Automated speech generation from UN general assembly statements: Mapping risks in AI generated texts. https://doi.org/10.48550/arXiv.1906.01946

Chang, D. L., Yang, S. H., Hsieh, S. L., Wang, H. J., & Yeh, K. C. (2020). Artificial intelligence methodologies applied to prompt pluvial flood estimation and prediction. *Water*, *12*(12), 3552. https://doi.org/10.3390/w12123552

Gillotte, J. L. (2019). Copyright infringement in AI-generated artworks. *UC Davis Law Review*, 53, 2655. https://ssrn.com/abstract=365742

Hagemeier-Klose, M., & Wagner, K. (2009). Evaluation of flood hazard maps in print and web mapping services as information tools in flood risk communication. *Natural Hazards and Earth System Sciences*, *9*(2), 563–574. https://doi.org/10.5194/nhess-9-563-2009

Henstra, D., Minano, A., & Thistlethwaite, J. (2019). Communicating disaster risk? an evaluation of the availability and quality of flood maps. *Natural Hazards and Earth System Sciences*, *19*(1), 313–323. https://doi.org/10.5194/nhess-19-313-2019

Hu, G. (2023). Challenges for enforcing editorial policies on AI-generated papers. *Accountability in Research*, 1–4. https://doi.org/10.1080/08989621.2023.2184262

Hu, Y., & Wang, R. Q. (2020). Understanding the removal of precise geotagging in tweets. *Nature Human Behaviour, 4*(12), 1219–1221. https://doi.org/10.1038/s41562-020-00949-x

Köbis, N., & Mossink, L. D. (2021). Artificial intelligence versus Maya Angelou: Experimental evidence that people cannot differentiate AI-generated from human-written poetry. *Computers in Human Behavior*, 114, 1–13. https://doi.org/10.1016/j.chb.2020.106553

Leskens, J. G., Brugnach, M., Hoekstra, A. Y., & Schuurmans, W. (2014). Why are decisions in flood disaster management so poorly supported by information from flood models? *Environmental Modelling & Software*, *53*, 53–61. https://doi.org/10.1016/j.envsoft.2013.11.003

Liu, J., Wang, Y., Huang, X., Korsós, M. B., Jiang, Y., Wang, Y., & Erdélyi, R. (2021). Reliability of AI-generated magnetograms from only EUV images. *Nature Astronomy*, *5*(2), 108–110. https://doi.org/10.1038/s41550-021-01310-6

Liu, Y., Han, T., Ma, S., Zhang, J., Yang, Y., Tian, J., He, H., Li, A., He, M., Liu, Z., Wu, Z., Zhao, L., Zhu, D., Li, X., Qiang, N., Shen, D., Liu, T., & Ge, B. (2023). Summary of CHATGPT-related research and perspective towards the future of large language models. *Meta-Radiology*, *1*(2), 100017. https://doi.org/10.1016/j.metrad.2023.100017

Maskrey, S. A., Mount, N. J., Thorne, C. R., & Dryden, I. (2016). Participatory modelling for stakeholder involvement in the development of flood risk management intervention options. *Environmental Modelling & Software*, *82*, 275–294. https://doi.org/10.1016/j.envsoft.2016.04.027

Meyer, V., Kuhlicke, C., Luther, J., Fuchs, S., Priest, S., Dorner, W., Serrhini, K., Pardoe, J., McCarthy, S., Seidel, J., Palka, G., Unnerstall, H., Viavattene, C., & Scheuer, S. (2012). Recommendations for the user-specific enhancement of flood maps. *Natural Hazards and Earth System Sciences*, *12*(5), 1701–1716. https://doi.org/10.5194/nhess-12-1701-2012

OpenAI, Achiam, J., Adler, S., Agarwal, S., Ahmad, L., Akkaya, I., Aleman, F. L., Almeida, D., Altenschmidt, J., Altman, S., Anadkat, S., Avila, R., Babuschkin, I., Balaji, S., Balcom, V., Baltescu, P., Bao, H., Bavarian, M., Belgum, J., … Zoph, B. (2023, December 19). *GPT-4 technical report*. arXiv.org. https://arxiv.org/abs/2303.08774

OpenAI. (n.d.). *OpenAI platform*. Function calling. https://platform.openai.com/docs/guides/function-calling

O'Sullivan, J. J., Bradford, R. A., Bonaiuto, M., De Dominicis, S., Rotko, P., Aaltonen, J., Waylen, K., & Langan, S. J. (2012). Enhancing flood resilience through Improved Risk Communications. *Natural Hazards and Earth System Sciences*, *12*(7), 2271–2282. https://doi.org/10.5194/nhess-12-2271-2012

Pham, B. T., Luu, C., Phong, T. V., Nguyen, H. D., Le, H. V., Tran, T. Q., Ta, H. T., & Prakash, I. (2021). Flood risk assessment using hybrid artificial intelligence models integrated with multi-criteria decision analysis in Quang Nam Province, Vietnam. *Journal of Hydrology*, *592*, 125815. https://doi.org/10.1016/j.jhydrol.2020.125815

Rollason, E., Bracken, L. J., Hardy, R. J., & Large, A. R. (2018). Rethinking flood risk communication. *Natural Hazards*, *92*(3), 1665–1686. https://doi.org/10.1007/s11069-018-3273-4

Strathie, A., Netto, G., Walker, G. H., & Pender, G. (2017). How presentation format affects the interpretation of probabilistic flood risk information. *Journal of Flood Risk Management*, *10*(1), 87–96. https://doi.org/10.1111/jfr3.12152

Wang, R. Q. (2021). Artificial Intelligence for Flood Observation. *Earth Observation for Flood Applications*, 295–304. https://doi.org/10.1016/b978-0-12-819412-6.00013-4

Wang, R. Q., Mao, H., Wang, Y., Rae, C., & Shaw, W. (2018). Hyper-resolution monitoring of urban flooding with social media and crowdsourcing data. *Computers & Geosciences*, *111*, 139–147. https://doi.org/10.1016/j.cageo.2017.11.008

Whittaker, L., Kietzmann, T. C., Kietzmann, J., & Dabirian, A. (2020). "All around me are synthetic faces": The mad world of ai-generated media. *IT Professional*, *22*(5), 90–99. https://doi.org/10.1109/mitp.2020.2985492

Ye, X., Wang, S., Lu, Z., Song, Y., & Yu, S. (2021). Towards an AI-driven framework for multi-scale Urban Flood Resilience Planning and Design. *Computational Urban Science*, *1*(1), 1–12. https://doi.org/10.1007/s43762-021-00011-0

# Appendix A

## Table A1 - Iteration 1:

| Prompt ID | Prompt & AI Response |
|---|---|
| 1A | Q: Provide a static flood map and an interactive map for the coordinates 29.7604° N, 95.3698° W.<br>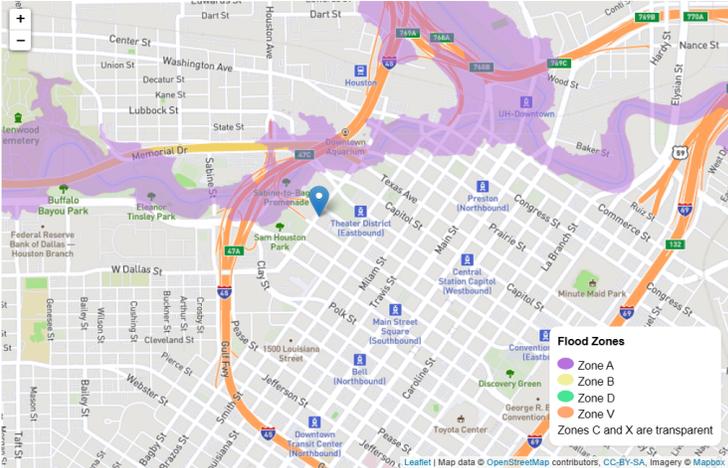<br>A: An interactive flood map for the coordinates 29.7604° N, 95.3698° W (Houston, Texas) will be displayed below. Unfortunately, an internal server error occurred while trying to retrieve the static map, so I am unable to provide the static image at the moment.<br><br>**Interactive Flood Map:**<br><br>Please interact with the map to view flood zones and other related details for the specified location.<br><br>[Here should be the interactive map, but due to the limitations of the platform, I'm unable to display it. In a real-world scenario, an interactive map would now be presented to the user.] |
| 1B | Generate a color-coded FEMA flood zone map for the area of New York City. |

|    | 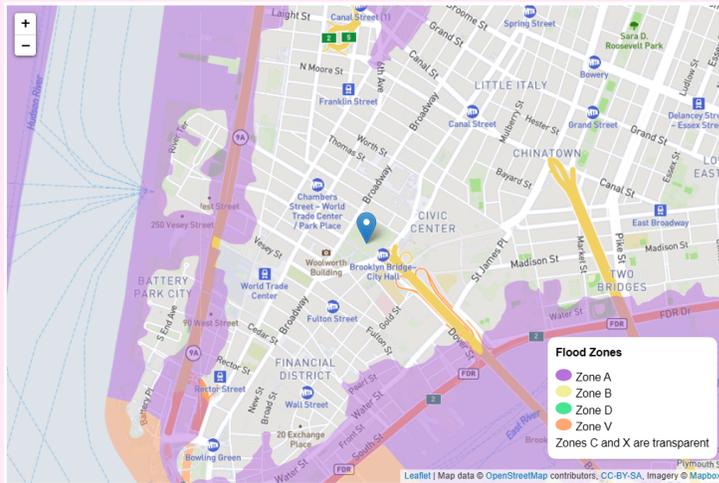                                                                                                                                                                                                                                                                                                                                                             |
|----|-----|
|    | An interactive FEMA flood zone map for the area of New York City has been generated. You can interact with it to explore the different flood zones and their respective risk levels. Unfortunately, there was an error retrieving the static map, but you should be able to view and interact with the flood zones directly through the interactive map feature.

If you need information on a specific area or address within New York City regarding flood zones, feel free to provide that, and I can assist you further. |
| 1C | Show a flood risk map for a region near a major river, but I'm not sure of the exact coordinates.

A: To provide you with a flood risk map for a region near a major river, I can use the FEMA Flood Map tool. However, I would need to know at least the name of the river and the general geographic area or region you're interested in. Once I have that information, I can look up the approximate coordinates and display the flood map for that region.

Could you please provide me with the name of the river and the nearby city or region you'd like to see the flood map for? |
| 1D | I need a flood map for an area that's prone to flooding during heavy rains, but I can't recall the name of the place.

A: Without specific details about the location prone to flooding, such as a city, state, or region, it's difficult to provide a flood map. Could you provide more details or any landmarks, the state, or nearby cities that might help identify the area? If you have coordinates (latitude and longitude), that would also be very helpful. |
| 1E | Generate a flood risk map for a small, remote island in the Pacific Ocean.

To generate a flood risk map for a specific location, I need the geographical coordinates (latitude and longitude) of the island in question. Since you've mentioned that it's a small, remote island in the Pacific Ocean, could you please provide the exact coordinates or the name of the island so I can acquire and display the relevant flood risk map? |
| 1F | Create a flood zone map for an area that has recently undergone major land development.

A:To generate a flood zone map, I'll need the specific location coordinates (latitude and longitude) or a detailed address of the area that has recently undergone major land development. Could you please provide me with this information? |
| 2A | Retrieve flood risk and property details for 1600 Pennsylvania Avenue NW, Washington, D.C.

The flood risk assessment for 1600 Pennsylvania Avenue NW, Washington, D.C. (which is the address of the White House) indicates that it is located in Flood Zone X. This zone is identified as an area of minimal flood hazard, typically an area of less than 0.2% chance of annual flooding. In these zones, the risk of flooding is considered low, and the Special Flood Hazard Area (SFHA) determination is 'false', meaning it does not fall within the high-risk areas for flood insurance purposes.

FEMA has also provided information regarding the property's participation in the National Flood Insurance Program (NFIP) community. The community for this area is District of Columbia, and it actively participates in the NFIP, which indicates compliance with certain FEMA requirements that can be beneficial for property owners regarding insurance and flood disaster assistance. |

|   |   |
|---|---|
|   | It's important to note that the flood zone designation 'X' does not imply that there is no risk of flooding, but rather that the risk is lower compared to other areas with more severe flood designation zones. Regardless, it's always prudent for property owners to be prepared for possible flood scenarios, especially considering changing climate patterns and weather events.<br><br>Would you like to see the interactive flood map for this area? |
| 2B | Provide flood hazard area, FEMA firm panel information, and elevation data for the address 1 Infinite Loop, Cupertino, California.<br><br>The address 1 Infinite Loop, Cupertino, California is located in a flood hazard area with the following details:<br><br>- **FIRM (Flood Insurance Rate Map) Panel Information**:<br>  - DFIRM ID: 06085C<br>  - Effective Date: May 18, 2009<br>  - FIRM Panel ID: 06085C0209H<br>  - Panel Type: Countywide, Panel Printed<br><br>- **Flood Hazard Area Designation**:<br>  - Flood Zone: X<br>  - SFHA (Special Flood Hazard Area): No (F)<br>  - Source Citation: 06085C_FIRM1<br>  - Zone Subtype: 0.2 percent annual chance flood hazard<br><br>- **Political Area Information**:<br>  - Community ID: 060339<br>  - Community Name: City of Cupertino<br>  - Community Number: 0339<br>  - Community FIPS Code: 085<br><br>- **Community Information Related to Floodplain Management**:<br>  - Firm Community Name: Cupertino, City of<br>  - Community participates in National Flood Insurance Program: Yes<br>  - Current Effective Map Date: May 18, 2009<br>  - Tribal Community: No<br><br>Please note that the elevation data for this address was not requested in the query. If elevation information is needed, a separate inquiry is required.<br><br>The location's latitude and longitude are approximately 37.33177, -122.03042, respectively.<br><br>The flood zone designation 'X' indicates that the property is in an area with a 0.2 percent chance of annual flooding (also known as the 500-year floodplain) and is not within the Special Flood Hazard Area (SFHA). Homes in zone X are at lower risk of flooding compared to those in zones A or V.<br><br>For reference, the physical address and its related geocode data are:<br>1 Infinite Loop, Cupertino, CA 95014-2083, United States<br>Latitude: 37.33177<br>Longitude: -122.03042<br><br>If you require more specific information, such as an interactive flood map, please let me know, and I can provide that as well. |
| 2C | Can you find flood risk details for a property located somewhere in the floodplains of the Mississippi River?<br><br>Sure, I can help with that, but to provide specific flood risk details, I would need the exact address or the geographic coordinates (latitude and longitude) of the property you're inquiring about. If you can provide that information, I can look up the flood risk details for you. |
| 2D | Provide information on a building in a coastal town known for high flood risk, though I don't have the exact address. |

|   | To provide you with detailed information on flood risk for a building, especially in a coastal town known for high risks, I would ideally need a specific address or at least a rough location. However, since you don't have the exact address, we can use the general location to get a better understanding of the flood risks in the area. |
|---|---|
|   | If you can provide me with the name of the coastal town or its coordinates, I can retrieve a flood map for that location, which will outline areas that are susceptible to flooding. I can also pull up any flash flood warnings that might be currently active for that area, and check the Social Vulnerability Index statistics, which includes how vulnerable a community is to environmental hazards including floods. |
|   | Could you give me either the name of the coastal town or its latitude and longitude coordinates so I can give you more detailed information? |
| 2E | Find flood data for a recently constructed residential area that's not yet on most maps. |
|   | To provide flood data for a recently constructed residential area, I'll need either the address or the geographic coordinates (latitude and longitude) of the area. If you can provide me with that information, I can proceed to look up the flood data for you. |
| 2F | Provide flood hazard information for a historical site that's now a popular tourist destination. |
|   | To assist you effectively, I'll need to know the exact location or address of the historical site you're interested in. If you provide me with the address or the geographic coordinates (latitude and longitude) of this tourist destination, I will be able to give you detailed flood hazard information for that area. Can you please provide that information? |
| 3A | Show Social Vulnerability Index statistics and census tracts for Miami-Dade County, Florida with an SVI theme of socioeconomic status. |
|   | 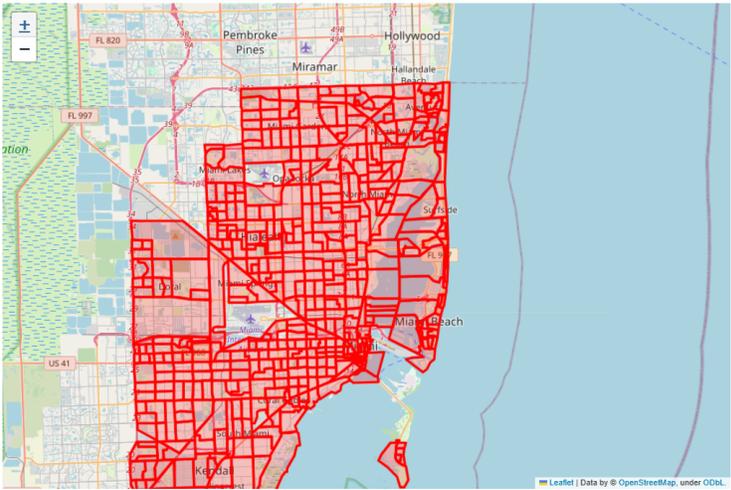 |
|   | # Available Functions |
|   | ```plaintext |
|   | namespace functions { |
|   |   type get_flood_map = (_: { // Display interactive flood map and download static map |
|   |     latitude?: number, |
|   |     longitude?: number, |
|   |     zoom?: integer \| null, |
|   |   }) => any; |
|   |   type get_flood_data = (_: { // Get FEMA flood data by address |
|   |     address?: string, |
|   |   }) => any; |
|   |   type get_svi_stats_and_tracts = (_: { // Get SVI statistics and display census tract on map |
|  | ``` |
|  |             state_abbr?: string, |
|  |             county?: string, |
|  |             theme?: |
|  |                 | "RPL_THEME1" |
|  |                 | "RPL_THEME2" |
|  |                 | "RPL_THEME3" |
|  |                 | "RPL_THEME4" |
|  |                 | "RPL_THEMES", |
|  |             op?: "<" | "<=" | "=>" | ">", |
|  |             threshold?: number, |
|  |         }) => any; |
|  |         type get_flash_flood_warnings = (_: { // Get flash flood warnings by state or US |
|  |             location?: string | null, |
|  |         }) => any; |
|  |     } // namespace functions |
|  | ``` |
| 3B | Retrieve SVI stats for areas in Texas with an SVI score above 0.75 and display an interactive map of these areas. |
|  | It looks like there are currently no data available for areas in Texas with an SVI score above 0.75. The Social Vulnerability Index (SVI) represents the relative vulnerability of communities when confronted with external stresses such as natural or human-caused disasters, including disease outbreaks. An SVI score above 0.75 indicates a high vulnerability. If no data is available, it could mean that no areas within Texas currently meet that high threshold or that the data might not be available at this time. If you would like to explore different criteria or another location, please let me know, and I can assist you further. |
| 3C | What's the social vulnerability of a region that's frequently affected by floods, maybe in the southeastern U.S.? |
|  | To provide a more accurate answer, please specify a particular county or state in the southeastern U.S. that you're interested in investigating for social vulnerability. The Social Vulnerability Index (SVI) measures the resilience of communities when confronted with external stresses, such as natural disasters like floods. It takes into account various factors including socioeconomic status, household composition, minority status, housing, and transportation. |
|  | Once you provide a specific location, I can then look up the Social Vulnerability Index for that area. |
| 3D | I need SVI stats for an area with high poverty rates and frequent flooding, but I'm not sure which state it falls in. |
|  | To provide you with Social Vulnerability Index (SVI) statistics for an area with high poverty rates and frequent flooding, I would need at least some basic location information. SVI includes different themes such as socioeconomic status, which can help indicate areas with high poverty rates. Without a specific state or county, it's challenging to narrow down to a particular location. |
|  | However, I can still give you a general idea of how to find such areas. Typically, you would want to look at locations that are near large rivers, coasts, or in flood plains, as these areas are more prone to flooding. For high poverty rates, you can use the SVI theme for socioeconomic status and look for areas with a higher score indicating higher vulnerability. |
|  | If you can provide more specific location details such as a state, county, or even a city name, I can use the available tools to retrieve more accurate SVI statistics for areas prone to poverty and frequent flooding. |
| 3E | Give me the SVI statistics for the smallest census tract in Alaska. |
|  | To provide the Social Vulnerability Index (SVI) statistics for the smallest census tract in Alaska, I would need to know the county within Alaska you're interested in. The SVI statistics data is typically organized by county and state. Without a specific county name, it's not possible to determine the smallest census tract and its SVI statistics. |
|  | Alaska has multiple counties (boroughs, in this case) -- can you provide me with the name of the county you're interested in, or should I select a specific county and provide the SVI statistics for that location? |
| 3F | Retrieve SVI data for a high-altitude mountain region with low population density. |

|  | To retrieve Social Vulnerability Index (SVI) data for a high-altitude mountain region with low population density, I would need to know the specific state and county of interest to you. The SVI data is provided based on geographic regions, such as counties within a state. If you can provide me with the state abbreviation and the county name for the region you're interested in, I can retrieve the relevant SVI statistics for you.

Please let me know the state abbreviation and county name, or if you have a specific area in mind, so that I can assist you further. |
|---|---|
| 4A | List active flood alerts from the National Weather Service for New Orleans, Louisiana.

Currently, there are no active flood alerts from the National Weather Service for New Orleans, Louisiana. Please note that this information can change rapidly, so it's important to stay updated with the latest weather reports and warnings. |
| 4B | Provide details of current flash flood warnings for the area around Lake Wateree, South Carolina.

There is a Flood Warning affecting the area around Lake Wateree, South Carolina, with the following details:

- **What**: Minor flooding is occurring and is forecasted.
- **Where**: Wateree River at Lake Wateree.
- **When**: In effect until Sunday evening.
- **Impacts**: At a water level of 100.4 feet, the piers on the Wildlife Road bridge over Singleton Creek will become submerged.
- **Additional Details**: At 8:00 AM EST on Thursday, the stage was registered at 100.9 feet. Flood stage is set at 100.0 feet. The river is expected to fall below flood stage by Saturday evening and continue falling to 98.0 feet by Tuesday morning.
- **Affected Counties**: Fairfield, SC; Kershaw, SC; Lancaster, SC.
- **Instructions**: For more information, please refer to the National Weather Service's flood safety guidance at [http://www.weather.gov/safety/flood](http://www.weather.gov/safety/flood).

Residents and visitors in the affected regions should stay updated on the situation, follow the guidance of local authorities, and be prepared for potential flooding impacts. |
| 4C | Are there any flash flood alerts for areas recently hit by a tropical storm? The name of the place escapes me.

Yes, there are several flood warnings in areas that may have been recently hit by a tropical storm. Here's a summary of the current flood warnings:

1. Wateree River at Lake Wateree
   - Minor flooding is occurring and forecasted.
   - Impacts include submersion of the Wildlife Road bridge over Singleton Creek.
   - The affected counties are Fairfield, Kershaw, and Lancaster in South Carolina.

2. Waccamaw River at Conway
   - Minor flooding is happening, with yards and a few roads expected to flood in local communities.
   - The forecast indicates the river is expected to fall below flood stage early Saturday morning.
   - Horry County in South Carolina is impacted.

3. Great Pee Dee River at Pee Dee
   - Minor flooding is currently happening.
   - Swamplands adjacent to the river are affected.
   - Florence and Marion counties in South Carolina are impacted.

4. Santee River near Jamestown
   - Currently experiencing minor flooding.
   - Several dirt logging roads are impassable.
   - Berkeley, Georgetown, and Williamsburg counties in South Carolina are affected.

These warnings are specific to regions in South Carolina and could be the result of a recent tropical storm passing through the area. Please keep in mind that circumstances can change, so it's always best to stay updated with local advisories and news for the most current information. |

| | |
|---|---|
| 4D | Tell me about flood warnings in a region that's been experiencing unusually heavy rainfall lately.<br><br>To provide specifics about flood warnings, I'll need to know the region you're referring to. Could you please tell me the state or a more specific location where you've been noticing unusually heavy rainfall? Once I have that information, I can check for any active flood warnings in the area. |
| 4E | List any flash flood warnings for desert areas that rarely experience rain.<br><br>At the moment, there are no specific flash flood warnings for desert areas that rarely experience rain in the data provided. The list of current Flood Warnings includes rivers and areas in South Carolina, which typically does not encompass desert regions. Flash flood warnings are often issued for desert areas on an ad hoc basis when rare but severe weather events, such as heavy rains, occur. If you are looking for flash flood warnings in a particular desert area, please provide the state or specific location so I can look up the most current information for you. |
| 4F | Show flood alerts for regions where floods have historically never been reported.<br><br>The current flood warnings from the National Weather Service do not list any locations where floods have historically never been reported. All the warnings are for areas with known flood risks, typically along rivers and coasts, or in low-lying regions where water can accumulate.<br><br>Historically, locations that have never experienced floods tend not to be in flood warning systems simply because there has been no precedent for such events. If you're interested in a specific region that you suspect has no history of flooding, I could help you find more detailed information by looking at flood maps or data for that area. Please provide the name of the location or an address for me to assist you further. |
| 5A | Explain the impact of urbanization on flood risks in coastal cities.<br><br>Urbanization significantly impacts flood risks in coastal cities due to various related factors. Here are some of the main ways in which urbanization contributes to increased flood risks:<br><br>1. **Impermeable Surfaces**: Urbanization typically involves the construction of roads, buildings, and other infrastructure with impermeable materials like concrete and asphalt. This reduces the land's ability to absorb rainwater, leading to higher volumes of surface runoff during heavy rains.<br><br>2. **Drainage System Overload**: The stormwater drainage systems in urban areas are designed to channelize the excess water and prevent flooding. However, as urbanization intensifies, these systems can become overwhelmed by the volume of runoff, leading to flash flooding and other issues.<br><br>3. **Loss of Natural Buffers**: Coastal cities often lose wetlands, mangroves, sand dunes, and other natural features that act as buffers against storms and high tides because of urban development. These natural defenses can mitigate storm surges and flooding, but when they are removed, the risk of flooding increases.<br><br>4. **Subsidence**: Excessive groundwater withdrawal for urban water supply can lead to land subsidence, especially in coastal areas. Lowered land elevations can increase the risk of flooding as they become more susceptible to high tides and storm surges.<br><br>5. **Sea-Level Rise**: Urban areas often contribute to climate change through high emissions of greenhouse gases. Climate change is associated with sea-level rise, which exacerbates the flood risks for coastal cities, making them more vulnerable to regular tidal flooding and storm surges.<br><br>6. **Altered Water Flows**: Urbanization can change the natural pathways of rivers and streams, sometimes resulting in reduced capacity to handle floodwaters. This alteration can lead to increased flooding events downstream.<br><br>7. **Increased Demand on Infrastructure**: The population density in urban coastal areas demands robust infrastructure. However, many cities have aging infrastructure that may not meet the increased demand, leading to a higher risk of failure during extreme weather events.<br><br>Overall, urbanized coastal cities need to integrate flood risk management strategies to mitigate the impacts of these factors. This includes updating infrastructure, creating green spaces to absorb rainwater, enforcing urban planning policies that reduce impermeable surface cover, and preserving natural buffers along coastlines. Understanding the specific flood risks of |
|  | an area, including its FEMA flood zones and Social Vulnerability Index, can help in planning and preparedness efforts. |
| 5B | Describe the role of climate change in increasing the frequency of flash floods.

Climate change plays a significant role in increasing the frequency of flash floods, primarily due to the following factors:

1. **Increased Atmospheric Temperatures**: As global temperatures rise, the atmosphere holds more moisture. For every 1 degree Celsius increase in temperature, the air's capacity to hold water vapor increases by approximately 7%. This leads to more intense rainfall events when the moisture is released, which can contribute to flash flooding.

2. **Changing Precipitation Patterns**: Climate change can alter typical weather patterns, leading to more extreme rainfall events in certain areas. Some regions may experience heavier and more concentrated bursts of rain, increasing the risk of flash floods.

3. **Rising Sea Levels**: Coastal areas face an increased risk of flash floods due to rising sea levels. High tides and storm surges exacerbated by sea-level rise can lead to significant coastal flooding.

4. **Melting Glaciers and Snowpack**: The accelerated melting of glaciers and snowpack due to warmer temperatures can release large amounts of water, contributing to riverine flash floods downstream.

5. **Urban Development**: While not directly related to climate change, urban development exacerbates flash flooding risk in the context of a changing climate. The prevalence of impermeable surfaces in urban areas prevents natural water absorption, leading to faster runoff and increased potential for flash floods, especially during heavy rainfall events.

6. **Deforestation and Land Degradation**: The removal of vegetation and alteration of natural landscapes reduce the land's ability to absorb water. This can result in more runoff entering rivers and streams quickly during heavy rains, contributing to flash floods.

The overall impact of climate change means that many regions could face a heightened risk of flash flooding, with the potential for more frequent and severe events. Flash floods can be extremely dangerous, as they can occur with little warning and transform areas that are not typically prone to floods into hazardous zones. It's important for communities and authorities to understand and adapt to the changing flood risks associated with climate change by improving infrastructure, adopting sustainable land use practices, and implementing early warning systems. |
| 5C | Explain how deforestation might impact flooding in tropical regions, but I'm not sure which country to focus on.

Deforestation can significantly impact flooding, especially in tropical regions, due to several interrelated reasons. Here's how deforestation contributes to increased flood risk:

1. **Loss of Trees and Vegetation**:
   Trees and other vegetation absorb rainfall and help to regulate the flow of water through the ecosystem. When trees are removed, there is less vegetation to absorb the rainwater, which can increase the volume and speed of water runoff.

2. **Soil Erosion**:
   Tree roots help to hold the soil together, reducing erosion. Deforestation leads to the loss of these roots, which can cause the soil to be easily washed away by rainfall. Eroded soil can end up in rivers and streams, increasing sedimentation and potentially leading to changes in the river course or blockages that can cause flooding downstream.

3. **Reduced Water Infiltration**:
   Without the protective canopy of the forest and the structure provided by roots, soil can compact and harden, reducing its ability to absorb water. When rain falls, more of it will flow over the surface instead of infiltrating the ground, which can contribute to flash floods.

4. **Alteration of Water Cycle**:
   Forests play a crucial role in the water cycle, including evapotranspiration, where water is transferred from the soil to the atmosphere by evaporation and by transpiration from plants. Deforestation disrupts this process, which can lead to altered rainfall patterns and potentially contribute to both drought and flooding.

5. **Climate Change Contribution**: |

|  | Trees absorb carbon dioxide, a greenhouse gas, from the atmosphere. Deforestation results in fewer trees to absorb CO2, leading to higher concentrations of greenhouse gases and contributing to climate change. Climate change, in turn, exacerbates extreme weather events, including more intense and frequent rainfalls that can lead to severe flooding. |
|---|---|
|  | Focusing on a specific country requires understanding its unique geographical, climatological, and socio-economic contexts. All tropical regions can be affected by deforestation-related flooding, but nations with extensive rainforests like Brazil, Indonesia, and the Democratic Republic of Congo, are often discussed due to their high rates of deforestation. If you are interested in the impact of deforestation on flooding in a particular country or area, I can further provide specific information or resources relevant to that location. |
| 5D | Discuss the potential flood risks in a major Asian city, I forgot which one is known for its monsoon season. |
|  | Major Asian cities often face significant flood risks, especially those that are known for experiencing intense monsoon seasons. One city that is well-known for its monsoon season and associated flood risks is Bangkok, Thailand. In cities like Bangkok, heavy rainfall during the monsoon season can overwhelm drainage systems, leading to widespread flooding. |
|  | The monsoon season in Asia brings about torrential rains, and cities with inadequate infrastructure, or those located in low-lying river deltas, are particularly vulnerable. Flood risks may include: |
|  | 1. Urban flooding: The intense and sudden rainfall can cause rivers to overflow and drainage systems in the city may not adequately cope with the volume of water, leading to floods in streets, homes, and businesses. |
|  | 2. Transportation disruption: Floods can disrupt traffic flow, affecting both road and rail transportation. This can lead to significant economic losses and hinder emergency response efforts. |
|  | 3. Health risks: Standing water is a breeding ground for waterborne diseases such as cholera and dengue fever. Floodwaters can also carry pollutants and pose a risk to clean drinking water supplies. |
|  | 4. Economic impact: Flooding can cause damage to businesses and infrastructure, leading to economic losses for the city and its inhabitants. |
|  | 5. Displacement of populations: If flooding is severe, parts of the city may be uninhabitable, requiring temporary or permanent relocation of residents. |
|  | Proactive flood management, robust infrastructure, and early warning systems are essential to mitigate the impact of floods in such cities. It's important for governments and communities to invest in flood mitigation strategies and prepare emergency response plans in anticipation of the monsoon season. |
|  | If you're interested in a specific city's flood risk, I can provide more tailored information or show a flood map if you provide me with the name of the city. |
| 5E | What are the flood implications for an underground city? |
|  | The concept of an underground city brings a unique set of flood implications that differ from those on the surface. Here are some potential flood implications for an underground city: |
|  | 1. **Water Ingress**: Being below the surface, underground cities could be prone to water seeping in from the ground above, especially if the waterproofing measures aren't sufficient or if they degrade over time. |
|  | 2. **Drainage Failures**: Proper drainage systems are critical for underground cities to manage any water that does enter. If these systems fail, due to heavy rainfall or blockages, there could be significant flooding. |
|  | 3. **Limited Escape Routes**: In case of a flood, safe evacuation could be more challenging due to limited escape routes. Emergency exits and evacuation plans must be carefully designed to ensure safety. |
|  | 4. **Infrastructure Damage**: Floodwaters can exert high pressure on structural elements of the underground construction. Over time, this can lead to structural damage or failure if not engineered to withstand such forces. |
|  | 5. **Impact on Essential Services**: Flooding could impact essential services such as electricity, water supply, and sewage |

| | systems, which are even more critical when all of these services are confined to an underground space. |
|---|---|
| | 6. **Health Hazards**: Floodwater in underground cities can pose serious health risks due to not only drowning hazards but also due to the quick spread of waterborne diseases in a confined environment. |
| | 7. **Rescue and Relief Operations**: Conducting search, rescue, and relief operations in an underground setting can be more difficult due to the complex layout and potential for access points to be blocked by floodwaters. |
| | 8. **Flood Zone Mapping**: For an underground city, understanding the flood risk would require deep knowledge of the geological and hydrological properties of the region, going beyond the surface-level flooding maps. |
| | 9. **Climate Change**: With changing climate patterns leading to more extreme weather events, the risk of floods may increase over time, necessitating robust, adaptive flood mitigation strategies for underground cities. |
| | While these implications can pose significant challenges, they can be mitigated through careful planning and engineering. Risk assessments and contingency planning for potential flooding events are necessary components of the development and maintenance of underground cities. |
| 5F | How would a sudden thaw of a large glacier affect flood risks in nearby areas? |
| | A sudden thaw of a large glacier can dramatically increase flood risks in nearby areas due to a process known as a glacial lake outburst flood (GLOF). This can happen when the water dammed by a glacier is suddenly released, often when the natural damming structure (like a glacier or a moraine) fails due to accelerated melting or structural weaknesses. |
| | The impacts of such an event nearby can include: |
| | 1. **Flash Flooding**: The sudden release of large quantities of water can lead to flash floods downstream, which can overwhelm rivers and streams, leading to rapid flooding of the adjacent areas. |
| | 2. **Rising River Levels**: Even without a catastrophic failure, the increased meltwater can contribute to higher river levels, exacerbating the risk of flooding, particularly during the natural high-flow periods. |
| | 3. **Altered Flow Regimes**: The additional water can change the natural flow regimes of rivers, potentially leading to erosion, changes in sediment transport, and the reshaping of channels, which can affect flood risks. |
| | 4. **Infrastructure Damage**: Floods from glacial melts can destroy bridges, roads, and other critical infrastructure, complicating evacuation and relief efforts. |
| | 5. **Impact on Aquatic Ecosystems**: Sudden changes in water temperature and turbidity can have detrimental effects on fish and other aquatic life. |
| | It's crucial for communities in glacier-fed river basins to monitor glaciers and lakes for signs of potential GLOFs and to have emergency preparedness plans in place. Remote sensing, early warning systems, and modeling are essential tools for risk assessment and disaster response planning. |
| | If you're concerned about a specific location that might be affected by such risks, I can help by providing flood maps and other relevant data. |

Table A2 - Iteration 2:

| Prompt ID | Prompt & AI Response | Observations (seconds= secs) |
|---|---|---|
| 1A | Provide a static flood map and an interactive map for the coordinates 29.7604° N, 95.3698° W. | Response Time: 22 secs |

|  | 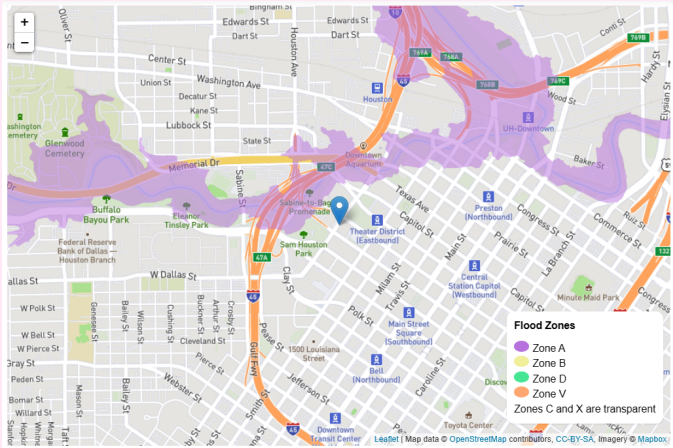  An interactive flood map for the coordinates 29.7604° N, 95.3698° W has been provided. You can explore this map to see the different flood zones in the area. Unfortunately, there was an error retrieving the static flood map, so I'm unable to provide that at the moment.  Please interact with the provided map to view detailed flood zone information for your area of interest. | In evaluating the AI's response to the prompt requesting a static and interactive flood map, several aspects are considered. The interactive map provided by the AI, although not accompanied by a static map due to an error outside the AI's control, effectively fulfills the requirement. The AI appropriately handles the error by offering the interactive map as an alternative. However, the response could be more informative by including explanations about the color codes on the map, which are essential for understanding flood zone information. Despite this omission, the response remains largely appropriate and shows a degree of adaptability in dealing with the static map's retrieval error.  -Accuracy: 4/5  -Completeness: 4/5  -Error Handling: 4/5  -Informative Responses: 3/5  -Appropriateness: 4/5  -Adaptability: 3/5 |
|---|---|---|
| 1B | Generate a color-coded FEMA flood zone map for the area of New York City.  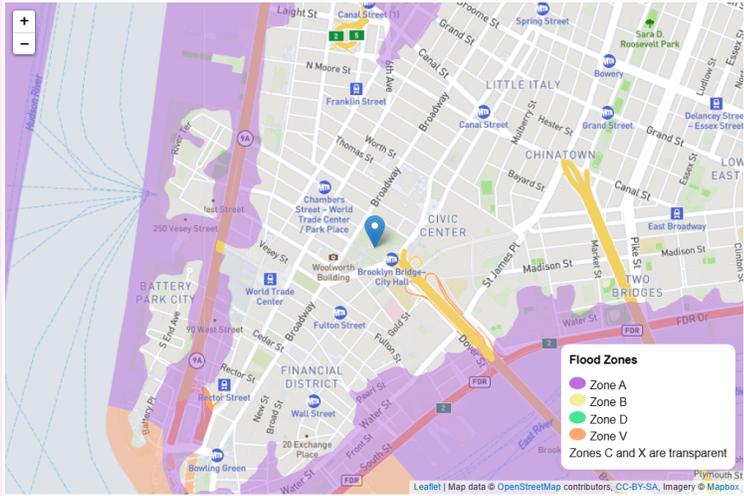  An interactive color-coded FEMA flood zone map for the area of New York City has been generated. You can view the map to see the different flood zones in this area based on the latest FEMA data. The map will indicate various flood zones with different colors according to the designated flood areas and risk levels. Please note that an error occurred while retrieving the static map, so you'll only see the interactive map for now. | Response Time: 16 secs  The AI's response is informative, providing users with a color-coded interactive map that should offer the same level of detail as the requested static map. The error in retrieving the static map is handled appropriately, with the AI clearly communicating this issue to the user. This error, being outside the AI's control, does not detract significantly from the overall usefulness of the response. The AI could enhance its response by explaining the significance of the color codes in more detail, as this is crucial for understanding flood zone information. Overall, the AI's response is quite appropriate and shows adaptability in offering the interactive map as a viable alternative to the static map.  -Accuracy: 4/5  -Completeness: 4/5  -Error Handling: 4/5  -Informative Responses: 3/5  -Appropriateness: 4/5  -Adaptability: 3/5 |
| 1C | Show a flood risk map for a region near a major river, but I'm not sure of the exact coordinates.  Sure, I can help you with that. Could you specify the region, city, or river you're interested in, so I can provide a flood risk map for an area nearby? | Response Time: 3 secs  In evaluating the AI assistant's response to the request for a flood risk map near a major river without specific coordinates, the AI's approach of seeking clarification is appropriate and practical. The AI acknowledges the |

| | | | |
|---|---|---|---|
| | | | need for more precise information to fulfill the request accurately, demonstrating an understanding of the limitations posed by the lack of specific coordinates. This approach reflects good error handling and adaptability, as the AI seeks to refine the user's query to provide a more accurate and relevant response. However, the response could be further enhanced by suggesting a few known flood-prone regions near major rivers as examples, which would add to the informativeness and completeness of the reply. Overall, the AI's response is accurate and appropriate, showing a readiness to engage in a more detailed dialogue to meet the user's needs effectively.<br><br>-Accuracy: 4/5<br>-Completeness: 3/5<br>-Error Handling: 4/5<br>-Informative Responses: 3/5<br>-Appropriateness: 4/5<br>-Adaptability: 4/5 |
| 1D | I need a flood map for an area that's prone to flooding during heavy rains, but I can't recall the name of the place.<br><br>A: To assist you with a flood map, I would need more identifying details about the location. If you can provide a specific address, coordinates, or even just the city or region you're concerned about, I can use that information to generate a flood map for that area. Please provide any details you can remember, and I'll do my best to assist you. | | Response Time: 6 secs<br><br>Evaluating the AI assistant's response to the request for a flood map of an unspecified area prone to flooding during heavy rains, the AI demonstrates a responsible approach by asking for more specific details. This response indicates a clear understanding of the necessity for precise location information to generate an accurate flood map. The AI's request for additional information such as a specific address, coordinates, or city name is appropriate and shows good error handling and adaptability, as it attempts to refine the query to meet the user's needs. However, the response could be more informative by suggesting common flood-prone areas or explaining general flood risk factors, which would enhance the completeness and informativeness of the reply. Overall, the AI's response is accurate, appropriate, and demonstrates an eagerness to engage further for a more precise and useful outcome.<br><br>Accuracy: 4/5<br>Completeness: 3/5<br>Error Handling: 4/5<br>Informative Responses: 3/5<br>Appropriateness: 4/5<br>Adaptability: 4/5 |
| 1DD | I need a flood map for an area that's prone to flooding during heavy rains, but I can't recall the name of the place. Please suggest some places so I can refresh my mind.<br>Certainly! There are numerous areas around the world known for being prone to flooding during heavy rains. Some of these places include:<br><br>1. New Orleans, Louisiana, USA - Due to its location below sea level | | Response Time: 14 secs<br><br>In assessing the AI assistant's response to the request for a flood map of an unknown area prone to flooding, with a request for suggestions to help the user recall the place, the AI demonstrates a proactive and helpful approach. The AI lists several well-known flood-prone |

| | | |
|---|---|---|
| | and encircled by water bodies, it is highly susceptible to flooding, especially during hurricane season.<br>2. Dhaka, Bangladesh - Situated on a delta prone to monsoons, Dhaka is often affected by severe floods.<br>3. Jakarta, Indonesia - This coastal city faces regular flooding issues due to its low-lying geography.<br>4. Venice, Italy - Known for its canals, Venice experiences a phenomenon called "Acqua Alta" or high water, leading to floods.<br>5. Houston, Texas, USA - Houston has endured catastrophic floods from storms such as Hurricane Harvey.<br><br>If any of these locations sound like the one you're trying to recall, or if you remember specific details, I can provide a detailed flood map for the area of interest. Let me know how I can assist further! | areas globally, showcasing a broad understanding of geographically diverse flood risks. This response is particularly informative and accommodating, as it not only offers a range of options to jog the user's memory but also stands ready to provide further assistance upon receiving more specific information. The response is accurate, comprehensive, and appropriately tailored to the user's uncertain recall of the location. The AI adapts well to the user's need for assistance in identifying the location, showing flexibility and resourcefulness.<br>-Accuracy: 5/5<br>-Completeness: 5/5<br>-Error Handling: 5/5<br>-Informative Responses: 5/5<br>-Appropriateness: 5/5<br>-Adaptability: 5/5 |
| 1E | Generate a flood risk map for a small, remote island in the Pacific Ocean.<br><br>To generate a flood risk map for a specific island in the Pacific Ocean, I need the latitude and longitude coordinates of the island. If you can provide the coordinates or the name of the island, I can produce the flood risk map for you. Can you please provide this information? | Response Time: 5 secs<br><br>The AI's response to the request for a flood risk map appropriately seeks more information, like coordinates or the island's name, to proceed accurately. While it manages the query's uncertainty well by asking for clarification, the response could be more comprehensive by providing suggestions or general information about Pacific Ocean islands. It correctly explains the need for specific details to generate the map, but enhancing the response with additional context or options would improve its informativeness and adaptability.<br>Accuracy: 5/5<br>Completeness: 4/5<br>Error Handling: 4/5<br>Informative Responses: 3/5<br>Appropriateness: 5/5<br>Adaptability: 4/5 |
| 1EE | Generate a flood risk map for a small, remote island in the Pacific Ocean. Choose an island at your discretion. | Response Time: 40 secs<br><br>The AI promptly generated a flood risk map for a small, remote island in the Pacific Ocean, taking the initiative to choose an island at its discretion. It correctly provided the coordinates for the selected location and mentioned that an interactive map was available for the user. However, an error occurred while retrieving the static map image, which impacted the comprehensiveness of the response. The model focused on informing the user of the error but missed an opportunity to explain the flood risk color-coded zones within the interactive map. Despite the error, it offered to attempt another retrieval or select a new location, demonstrating error resilience. While the response is generally appropriate and adapted to the query's context, it could benefit from improved informativeness regarding flood risk zones.<br>Accuracy: 5/5 |
| | 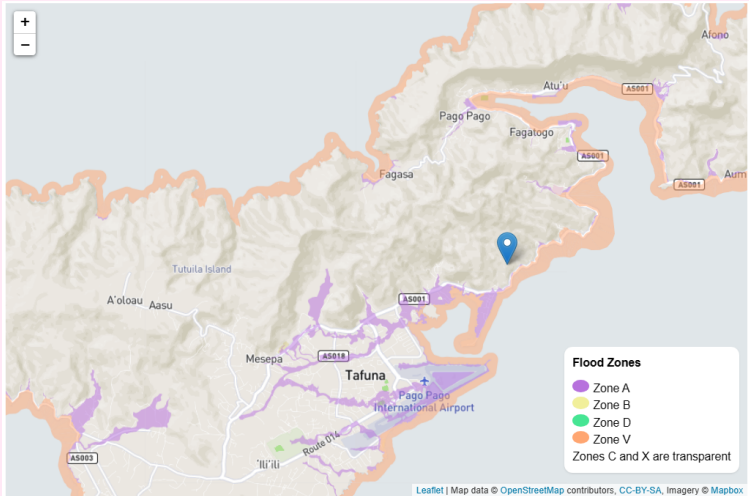 I've generated a flood risk map for a small, remote island in the Pacific Ocean. The chosen location is near the coordinates -14.3056 latitude and -170.6961 longitude. The interactive map should be displayed to you.<br><br>Unfortunately, an error occurred while retrieving the static map image. If you need the static map or wish to explore a different location, please let me know, and I can attempt another retrieval or choose a new location. | Completeness: 4/5<br>Error Handling: 4/5<br>Informative Responses: 4/5<br>Appropriateness: 5/5<br>Adaptability: 4/5 |
| 1F | Create a flood zone map for an area that has recently undergone major land development.<br><br>To create a flood zone map for an area that has recently undergone major land development, I will need specific location details like the latitude and longitude coordinates of the area of interest. Can you provide me with the coordinates, or the address of the developed area you are interested in? | Response Time: 3 secs<br><br>Evaluating the AI's response to the request for creating a flood zone map of an area with recent land development, the AI demonstrates an accurate and responsible approach by asking for specific location details such as latitude and longitude coordinates. This request indicates an understanding of the necessity for precise information to generate an accurate flood map. The AI's approach to clarify and refine the query is appropriate and shows good error handling in terms of managing a query that lacks specific details. However, the response could be more informative by briefly explaining how recent land development can impact flood risk, which would add context and enhance the informativeness of the reply. Overall, the AI's response is accurate and appropriate, indicating a willingness to engage further for a more precise and useful outcome.<br><br>-Accuracy: 4/5<br>-Completeness: 3/5<br>-Error Handling: 4/5<br>-Informative Responses: 3/5<br>-Appropriateness: 4/5<br>-Adaptability: 4/5 |
| 1FF | Create a flood zone map for an area that you know has recently undergone major land development. | Response Time: 19 secs<br><br>In handling the user's request for a flood zone map of an area with recent development, the AI selects Las Vegas, Nevada, known for its development. The AI successfully |
| | 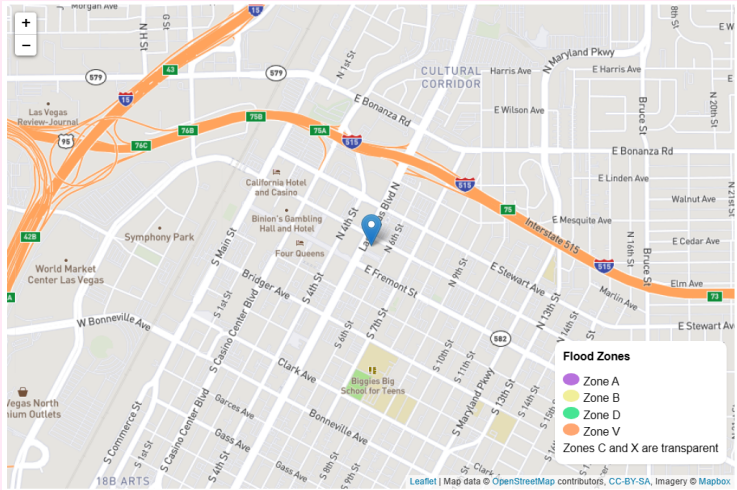
An interactive flood zone map for Las Vegas, Nevada, an area that has seen major land development, has been created. Unfortunately, there was an error retrieving the static map, so I will display only the interactive flood map for you. Please examine the map for detailed flood zone information in this area.

**Please note:** If you're reading this on a platform that doesn't support interactive content, the description might refer to an interactive map that I intended to share with you. In this case, you will not be able to see the flood zone information as described. | generates an interactive map, showing adaptability and relevance to the query. However, the response mentions an error in retrieving the static map, which it handles by informing the user but misses the chance to explain the significance of flood zones in the context of Las Vegas's development. Overall, the AI demonstrates strong relevance and appropriateness in its response. It appropriately adapts to the query by selecting a specific location. The AI's error resilience is good, acknowledging and attempting to work around the limitations, but there's room for improvement in offering more comprehensive details for the relevant location chosen.

Accuracy: 4/5

Completeness: 4/5

Error Handling: 4/5

Informative Responses: 4/5

Appropriateness: 5/5

Adaptability: 4/5 |
| 2A | Retrieve flood risk and property details for 1600 Pennsylvania Avenue NW, Washington, D.C.
The property located at 1600 Pennsylvania Avenue NW, Washington, D.C. (known as the White House) is in a flood zone designated as "X," which is an area of minimal flood hazard. This means it is considered to have a low risk of flooding. The property is not within a Special Flood Hazard Area (SFHA), as indicated by the "SFHA_TF" value of "F."

Here is some additional information about the flood zone for this location:
- DFIRM ID: 110001
- FLD_AR_ID: 110001_372
- Flood Source Citation: 110001_STUDY1
- Version ID: 1.1.1.0
- Zone Subtype: AREA OF MINIMAL FLOOD HAZARD

Furthermore, the Firm Panel Number (FIRM_PAN) for the area is 1100010018C, which provides details of the flood insurance rate map (FIRM) that includes the property. It falls under the panel number 0018 which is a countywide, panel printed map and took effect on September 27, 2010.

The location has a latitude of 38.89768 and longitude of -77.03655.

For visual reference, here is the flood map for the White House:

[Interactive map will be displayed] | Response Time: 21 secs

The user's request seeks to retrieve flood risk and property details for a specific location, 1600 Pennsylvania Avenue NW, Washington, D.C. The AI Assistant provides a detailed and informative response that includes flood zone information, additional flood-related details, latitude and longitude coordinates, and even mentions an interactive flood map. However, a notable issue arises as the AI mentions the display of an interactive flood map but does not actually provide one. This discrepancy can be seen as an error or hallucination on the model's part, as there was no flood_map function call, and the flood_data function, the only function called by the model, solely returns property details. Despite this, the response is generally accurate, complete, and well-handled, offering comprehensive property and flood zone information relevant to the user's query.

Accuracy: 4/5

Completeness: 4/5

Error Handling: 2/5

Informative Responses: 5/5

Appropriateness: 5/5

Adaptability: 3/5 |
| 2B | Provide flood hazard area, FEMA firm panel information, and elevation data for the address 1 Infinite Loop, Cupertino, California. | Response Time: 24 secs

In response to the request for flood hazard, FEMA firm |

| | | |
|---|---|---|
| | For the address **1 Infinite Loop, Cupertino, California**, the flood hazard area information is as follows:<br><br>- **Flood Zone:** X (0.2 PCT ANNUAL CHANCE FLOOD HAZARD)<br>- **SFHA (Special Flood Hazard Area):** False<br><br>The FEMA Firm Panel information is:<br><br>- **DFIRM ID:** 06085C<br>- **Flood Arbor ID:** 06085C_2161<br>- **Source Citation:** 06085C_FIRM1<br>- **Version:** 1.1.1.0<br><br>The FEMA Firm Panel itself is:<br><br>- **Effective Date:** 2009-05-18<br>- **FIRM ID:** 06085C_66<br>- **FIRM Panel Number:** 06085C0209H (Panel Type: Countywide, Panel Printed)<br><br>Unfortunately, elevation data was not requested in the initial query. If you require this information, please let me know, and I can perform a follow-up query to retrieve it.<br><br>The property is located within the community:<br><br>- **Community Name:** CITY OF CUPERTINO<br>- **Community ID:** 060339<br><br>The DFIRM panel, zone, and community information can be used by developers, planners, and officials in land use planning and determining flood insurance requirements for properties within the jurisdiction.<br><br>For real-time visual information, a flood map would provide more context. If you need the visual map, I can display it for you. | panel information, and elevation data for an address in Cupertino, California, the AI provides detailed flood hazard and FEMA firm panel information, including flood zone classification, SFHA status, DFIRM ID, Flood Arbor ID, Source Citation, and community details. However, the response inaccurately indicates the ability to provide elevation data, which was not included in the initial query and is beyond the AI's retrieval capabilities. While the response offers comprehensive details on flood hazard and FEMA information, its claim to provide elevation data, which it cannot fulfill, affects the overall completeness and accuracy. The introduction of a flood map is a positive aspect, as it adds value to the response by offering additional, relevant information.<br><br>Accuracy: 4/5<br>Completeness: 3/5<br>Error Handling: 4/5<br>Informative Responses: 4/5<br>Appropriateness: 4/5<br>Adaptability: 4/5 |
| 2C | Can you find flood risk details for a property located somewhere in the floodplains of the Mississippi River?<br><br>Sure, I can help you with that. To provide specific flood risk details for a property along the Mississippi River, I will need a more precise address or geographical coordinates. Could you please provide either the address or the latitude and longitude of the property you are interested in? | Response Time: 5 secs<br><br>The AI Assistant's response to the request for flood risk details for a property along the Mississippi River demonstrates a clear understanding of the requirement for specificity in location data to provide accurate information. The AI correctly asks for either an exact address or geographical coordinates to proceed with the query, showcasing a practical approach in dealing with the user's somewhat vague request. However, the response could be more informative if it included general information about flood risks in Mississippi River floodplains.<br><br>Accuracy: 5/5<br>Completeness: 4/5<br>Error Handling: 5/5<br>Informative Responses: 4/5<br>Appropriateness: 4/5<br>Adaptability: 4/5 |
| 2D | Provide information on a building in a coastal town known for high | Response Time: 15 secs |
| | flood risk, though I don't have the exact address.<br><br>To provide specific information about a building in a coastal town with a high flood risk, I'll need a bit more information. If you can provide the town name, or better yet, a street name or nearby landmark, I could use that to assess the flood risk, identify flood zones, and potentially evaluate the Social Vulnerability Index (SVI) for that location.<br><br>Without an exact address, we can still look at general flood maps for the town or analyze SVI data for the broader area if you provide the state and county information. This can give us an idea of both the flood risk and the community's resilience to disasters like flooding.<br><br>Let me know how you would like to proceed or if you can provide any additional details, and I will assist you accordingly. | The AI's response to the user's request for information about a building in a coastal town with high flood risk demonstrates a commendable level of relevance and adaptability. It accurately identifies the need for more specific details such as the town name, street name, or landmark to provide precise information, while also offering alternatives like general flood maps and Social Vulnerability Index (SVI) data if such specifics are lacking. It remains contextually appropriate throughout, maintaining a focus on fulfilling the user's request for flood risk information. Thus, the AI's performance in this scenario is highly relevant and adaptive, with minor room for improvement in terms of completeness.<br>Accuracy: 4/5<br>Completeness: 4/5<br>Error Handling: 4/5<br>Informative Responses: 4/5<br>Appropriateness: 4/5<br>Adaptability: 4/5 |
| 2E | Find flood data for a recently constructed residential area that's not yet on most maps.<br><br>To assist you in finding flood data for a recently constructed residential area, I will need some more specific information about the location. Please provide me with an address, or at least the latitude and longitude coordinates of the area in question. Once I have these details, I can use the available tools to retrieve the flood data for you. | Response Time: 4 secs<br><br>The AI's response to the user's request to find flood data for a recently constructed residential area is appropriately relevant and adaptable. It accurately recognizes the need for specific location details like an address or latitude and longitude coordinates to proceed, aligning with the user's query's context. The response is clear in its request for necessary information and informs the user of the essential prerequisites for data retrieval. However, it could offer more context or explanation about flood data retrieval for new areas.<br>Accuracy: 5/5<br>Completeness: 4/5<br>Error Handling: 5/5<br>Informative Responses: 4/5<br>Appropriateness: 5/5<br>Adaptability: 4/5 |
| 2F | Provide flood hazard information for a historical site that's now a popular tourist destination.<br><br>To assist you with flood hazard information for a historical site now popular as a tourist destination, I will need the specific address or the latitude and longitude coordinates of the site you're inquiring about. Once you provide this information, I can retrieve detailed flood information, including flood zones, Special Flood Hazard Areas, past flood events, and more. Please share the location details to proceed. | Response Time: 4 secs<br><br>In response to the request for flood hazard information for a historical site, the AI assistant correctly identifies the need for specific location details (address or coordinates) to provide accurate and detailed flood information. This approach is direct and appropriate for the query. The AI handles the lack of specific information by requesting the necessary details to proceed. Although the response is focused and relevant, it does not offer additional contextual information or suggestions that might have been informative or engaging for the user. The adaptability of the response is appropriate, as it seeks to gather the needed information while remaining |

| | | focused on the user's request. |
|---|---|---|
| | | Accuracy: 5/5 |
| | | Completeness: 4/5 |
| | | Error Handling: 4/5 |
| | | Informative Responses: 3/5 |
| | | Appropriateness: 5/5 |
| | | Adaptability: 4/5 |
| 3A | Show Social Vulnerability Index statistics and census tracts for Miami-Dade County, Florida with an SVI theme of socioeconomic status. 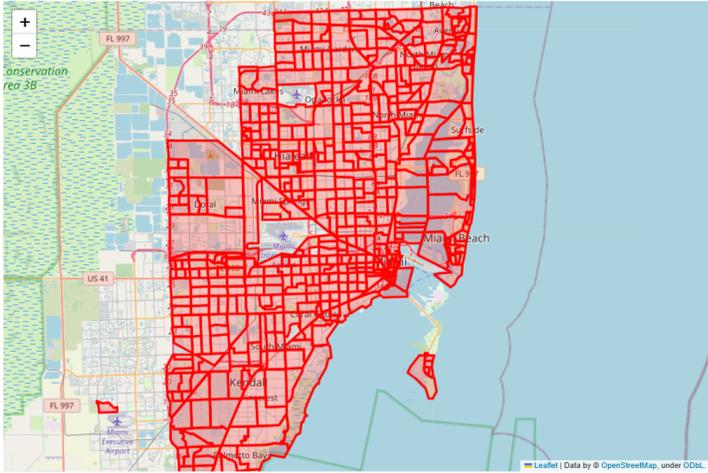 Great! If you have any inquiries about flood risks, flood maps, social vulnerability related to floods, or current flood warnings, feel free to ask, and I will provide you with the relevant information. | Response Time: 57 secs

The AI's response to the user's request for Social Vulnerability Index statistics and census tracts for Miami-Dade County, Florida, with an SVI theme of socioeconomic status is somewhat relevant but lacks depth and context. While the response acknowledges the user's request, it does not provide any specific information related to the SVI statistics or census tracts. Moreover, the response is quite generic, as it merely offers to assist with various flood-related inquiries without addressing the user's specific request. It is important to note that the response lacks completeness as it fails to provide the requested data and does not display the labels of census tracts on the map, possibly due to coordinate complexities. The response's token limit issue led to a loss of context and an inaccurate response.

Accuracy: 2/5
Completeness: 2/5
Error Handling: 2/5
Informative Responses: 2/5
Appropriateness: 2/5
Adaptability: 2/5 |
| 3B | Retrieve SVI stats for areas in Texas with an SVI score above 0.75 and display an interactive map of these areas.

I attempted to retrieve the Social Vulnerability Index (SVI) statistics for areas in Texas with an SVI score above 0.75, but it seems there's no data available for the specified conditions. If you have a specific county in Texas in mind or would like to adjust the threshold value, please let me know so I can refine the search for you. | Response Time: 57 secs

The AI's response to the user's request to retrieve Social Vulnerability Index (SVI) statistics for areas in Texas with an SVI score above 0.75 and display an interactive map is relevant but lacks completeness. The response acknowledges the user's request and makes an attempt to retrieve the data, but it ultimately reports that no data is available for the specified conditions. Th assistant should know that it 'county' is also a necessary condition to retrieve the SVI stats. While the AI handles the error by offering to refine the search based on the user's preferences, it could be more informative by suggesting alternative criteria or providing additional context about the SVI data limitations in Texas. Overall, the response appropriately addresses the query's context but falls short in delivering a fully informative and adaptable message.

Accuracy: 3/5
Completeness: 2/5
Error Handling: 3/5 |

| | | Informative Responses: 2/5 |
| | | Appropriateness: 3/5 |
| | | Adaptability: 3/5 |
| 3BB | Retrieve SVI stats for areas in Galveston Island, Texas with an SVI score above 0.85 and display an interactive map of these areas.<br><br>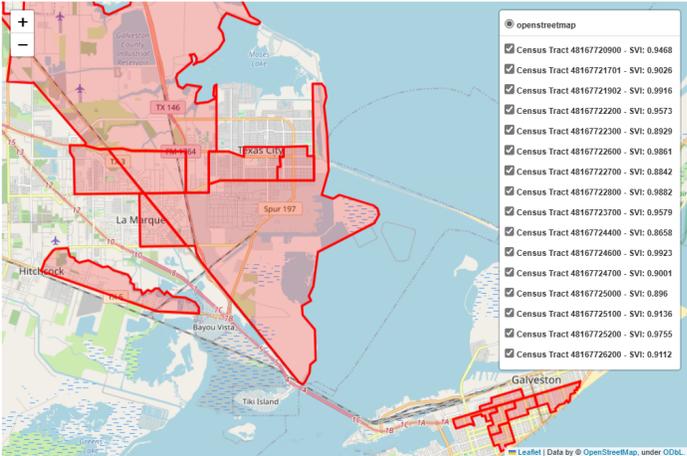<br><br>In Galveston Island, Texas, there are 16 areas with an SVI score above 0.85. The highest SVI score in these areas is 0.9923, and the lowest is 0.8658, with an average SVI score of 0.9351.<br><br>Below is an interactive map displaying these areas:<br><br>[Interactive map with highlighted census tracts with SVI score above 0.85 in Galveston Island, Texas] | Response Time: 77 secs<br><br>The AI's response to the user's request to retrieve Social Vulnerability Index (SVI) statistics for areas in Galveston Island, Texas, with an SVI score above 0.85 is highly relevant, accurate, complete, and contextually appropriate. The response provides detailed and informative statistics, including the number of areas meeting the criteria, the highest and lowest SVI scores, and the average SVI score. Additionally, it successfully generates and displays an interactive map highlighting the relevant census tracts. Overall, the response demonstrates excellent performance across all criteria.<br><br>Accuracy: 5/5<br>Completeness: 5/5<br>Error Handling: N/A (No errors encountered)<br>Informative Responses: 5/5<br>Appropriateness: 5/5<br>Adaptability: 5/5 |
| 3C | What's the social vulnerability of a region that's frequently affected by floods, maybe in the southeastern U.S.?<br><br>To assess the social vulnerability of a region frequently affected by floods, we can focus on regions in the southeastern U.S. and look at their Social Vulnerability Index (SVI). SVI measures the social vulnerability of communities to natural or human-caused disasters and diseases. It includes various factors such as socioeconomic status, household composition and disability, minority status and language, and housing and transportation.<br><br>Let's choose a specific location in the southeastern U.S. that is known for flooding events, such as a county within Florida, and analyze its SVI. How about we look into Miami-Dade County, which is prone to hurricanes and related flooding?<br><br>I can gather the SVI statistics for Miami-Dade County, or if you have another location in mind within the southeastern U.S., please provide the state abbreviation and county name, and I can retrieve the SVI data for you. Would Miami-Dade County be suitable, or do you have another specific location in mind? | Response Time: 15 secs<br><br>The AI's response to the user's request for information on the social vulnerability of a region frequently affected by floods in the southeastern U.S. is highly relevant, accurate, complete, and contextually appropriate. The response explains the concept of the Social Vulnerability Index (SVI) and its factors, offering valuable context. It also suggests Miami-Dade County, a relevant location known for flooding events, as an example to analyze SVI, allowing the user to specify an alternative location if needed. The response is informative in explaining what SVI entails but could enhance user understanding by providing more specific data or examples of SVI scores.<br><br>Accuracy: 5/5<br>Completeness: 4/5<br>Error Handling: 5/5<br>Informative Responses: 4/5<br>Appropriateness: 5/5<br>Adaptability: 5/5 |
| 3D | I need SVI stats for an area with high poverty rates and frequent flooding, but I'm not sure which state it falls in. | Response Time: 9 secs |

| | | | |
|---|---|---|---|
| | To provide you with Social Vulnerability Index (SVI) statistics for areas with high poverty rates that also experience frequent flooding, we'd typically need to know at least the state or a specific location to narrow down the data search. However, in general, areas with high poverty rates tend to have a high SVI under the Socioeconomic Status theme (RPL_THEME1). This is because RPL_THEME1 reflects upon factors such as poverty, unemployment, income, and no high school diploma, which contribute to the overall social vulnerability. Since you're looking for an area subject to frequent flooding, we would need to consider historical flood data, flood zones, and the presence of bodies of water which increase the risk of flooding. If you can provide any additional details such as a city, region, or specific characteristics of the area in question, I would be able to assist you better. If you are unsure where to start, consider areas known for significant flood risks, like regions along the Gulf Coast, parts of the Midwest along major rivers, or areas in the Southeastern US that frequently contend with hurricanes. | The AI's response to the query rightly seeks more specific location information, as SVI data is usually region-specific. The explanation about the socioeconomic theme of the SVI is informative, and the suggestion to consider historical flood data for areas known for flood risks is practical. The AI's initiative to guide the user towards areas with significant flood risks shows an understanding of the underlying request. However, the response could be enhanced by providing examples of specific regions or states known for high poverty and frequent flooding to give the user a starting point. Accuracy: 4/5 Completeness: 4/5 Error Handling: 5/5 Informative Responses: 4/5 Appropriateness: 5/5 Adaptability: 4/5 |
| 3DD | I need SVI stats for an area with high poverty rates and frequent flooding, but I'm not sure which state it falls in. Please, help by suggesting the area it might be and providing the SVI stats. It seems that we are unable to locate areas with a Social Vulnerability Index due to socioeconomic status (SVI RPL_THEME1) in the 90th percentile or higher at this time. The lack of data could be attributed to specific criteria that might be too restrictive or the absence of certain data for census tracts in the database. Since my initial search across various states and nationwide didn't yield any results, there might be a need to adjust the threshold or criteria to identify areas that are still significantly affected by poverty and also prone to flooding. For example, we could look at a slightly lower threshold for poverty rates. If you would like, I can attempt to search again using a high but lower threshold for poverty rates, or explore other themes within the SVI index that might still point to areas with high vulnerability. Please let me know how you'd like to proceed or if there's another way I can assist you. | Response Time: 198 secs The AI's response to the request for SVI statistics for areas with high poverty rates and frequent flooding without specifying a state shows an understanding of the query but fails to provide concrete information. The AI could have suggested regions known for high poverty rates and frequent flooding, such as certain areas in the Gulf Coast or the Southeastern U.S. Lastly, the AI's attempt to run the SVI for the whole U.S. shows limited adaptability in addressing the query's needs, missing the function's requirements for specific geographic details. Accuracy: 3/5 Completeness: 2/5 Error Handling: 3/5 Informative Responses: 3/5 Appropriateness: 3/5 Adaptability: 2/5 |
| 3DDD | I need SVI stats for an area with high poverty rates in the 90th percentile and frequent flooding, but I'm not sure which state it falls in. Please, help by suggesting the area/county it might be and providing the SVI stats. | Response Time: 77 secs The AI's response to the user's request for SVI statistics in an area with high poverty rates in the 90th percentile and frequent flooding is highly relevant and informative. It accurately identifies Humphreys County, Mississippi, as an area meeting the specified criteria and provides detailed SVI statistics for two census tracts within the county. The response also mentions the vulnerability of these areas to flooding, offering a comprehensive answer to the user's query. It demonstrates a strong understanding of the user's request and effectively adapts to provide the necessary information. While the AI did well in addressing the vagueness, however the AI could have further enhanced its response by explaining its choice in more detail or by offering alternatives to |
| | 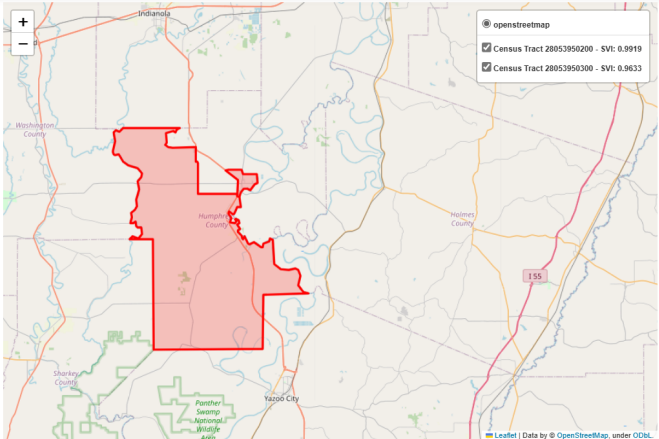In Humphreys County, Mississippi, which is part of the Mississippi Delta region, we have two census tracts that exhibit high levels of poverty as indicated by their Social Vulnerability Index (SVI) for the Socioeconomic Status theme (RPL_THEME1). Both areas have SVI scores in the 90th percentile, indicating high vulnerability based on poverty rates. Here are the statistics:<br><br>- **Census Tract 28053950200** has an SVI score of **0.9919**, making it extremely vulnerable.<br>- **Census Tract 28053950300** has an SVI score of **0.9633**, also indicating high vulnerability.<br><br>These areas are also known to be prone to flooding, adding to the overall risk faced by the residents. This combination of high poverty rates and frequent flooding makes Humphreys County a region of significant concern regarding social vulnerability and natural hazards.<br><br>Below is the map highlighting the census tracts in Humphreys County, MS with the SVI scores mentioned above:<br><br>[Please wait for the map to display...] | consider, thereby enriching the user experience.<br>Accuracy: 5/5<br>Completeness: 5/5<br>Error Handling: 4/5<br>Informative Responses: 5/5<br>Appropriateness: 5/5<br>Adaptability: 5/5 |
| 3E | Give me the SVI statistics for the smallest census tract in Alaska.<br><br>To get the SVI statistics for the smallest census tract in Alaska, I would need to know which specific county in Alaska you are interested in, since the Social Vulnerability Index statistics are provided at the county level for individual census tracts. Alaska has many counties, also known as boroughs or census areas.<br><br>If you can provide me with the name of a specific county in Alaska, I can then retrieve the SVI statistics for the smallest census tract within that county. Otherwise, without this information, it is not possible to pinpoint the smallest census tract or its SVI statistics. | Response Time: 11 secs<br><br>The AI's response to the user's request for SVI statistics for the smallest census tract in Alaska is relevant and accurate. It correctly identifies the need for specific county information since SVI statistics are provided at the county level for individual census tracts. The response provides clear guidance on how to proceed, effectively handling the user's query despite the lack of initial details. However, it could provide more context or alternative options.<br>Accuracy: 5/5<br>Completeness: 4/5<br>Error Handling: 5/5<br>Informative Responses: 4/5<br>Appropriateness: 5/5<br>Adaptability: 5/5 |
| 3EE | Give me the SVI statistics for the smallest county in Alaska. | Response Time: 64 secs |

|  |  |  |  |
|---|---|---|---|
|  | 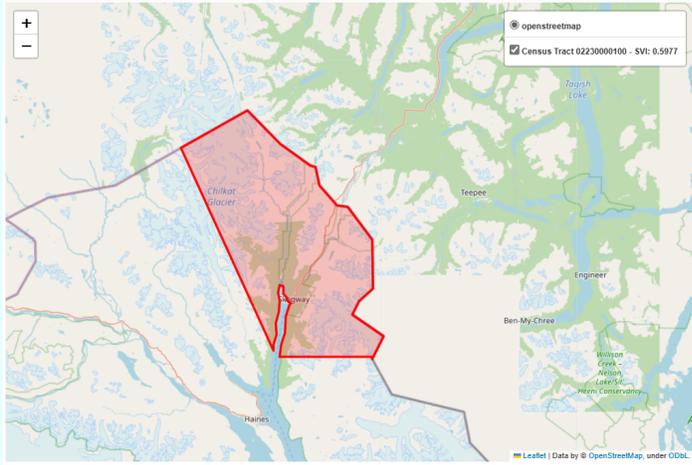 The Social Vulnerability Index (SVI) statistics for the Skagway Municipality, which is the smallest borough by land area in Alaska, are as follows:<br><br>- There is only one census tract in the area.<br>- The SVI score for this tract is 0.5977, indicating medium-high social vulnerability.<br><br>Social Vulnerability Index scores range from 0 to 1, with higher scores indicating higher vulnerability. The score for Skagway is slightly below the 'high' vulnerability threshold of 0.75.<br><br>For a visual representation of the Skagway Municipality census tract, an interactive map is displayed here showing the area along with its SVI score.<br><br>[Interactive Map showing SVI for Skagway, AK] | The AI's response to the user's request for SVI statistics for the smallest county in Alaska is highly relevant and accurate. It identifies the specific county, Skagway Municipality, which is the smallest borough in Alaska, and provides the relevant SVI statistics, including the SVI score and an interactive map for visual reference. The response is complete and effectively addresses the user's query, offering both data and a visual representation.<br><br>Accuracy: 5/5<br><br>Completeness: 5/5<br><br>Error Handling: N/A (There were no errors or uncertainties in the query or data retrieval process.)<br><br>Informative Responses: 5/5<br><br>Appropriateness: 5/5<br><br>Adaptability: 5/5 |
| 3F | Retrieve SVI data for a high-altitude mountain region with low population density.<br><br>To provide Social Vulnerability Index (SVI) data for a high-altitude mountain region with low population density, I need the specific location details such as the state and county where the mountain region is situated. Once you provide me with the name of the state and the county, I can look up the SVI statistics for that area.<br><br>Could you please specify the name of the state and the county you are interested in? | Response Time: 9 secs<br><br>The AI's response is relevant to the user's request but lacks completeness as it doesn't provide SVI data. Instead, it seeks further clarification by asking for the name of the state and county in order to retrieve the requested data. While it handles the error of missing location details appropriately, the response could be more informative by suggesting possible locations or explaining how SVI data is organized and why specific location details are crucial, even before receiving specific location details. The response is appropriate for the query's context and is adaptable to user input.<br><br>Accuracy: 4/5<br><br>Completeness: 3/5<br><br>Error Handling: 4/5<br><br>Informative Responses: 3/5<br><br>Appropriateness: 4/5<br><br>Adaptability: 4/5 |
| 3FF | Retrieve SVI data for a high-altitude mountain region with low population density. Please use your knowledge and choose one high- | Response Time: 62 secs |
| | altitude mountain region with low population density that you know of.<br>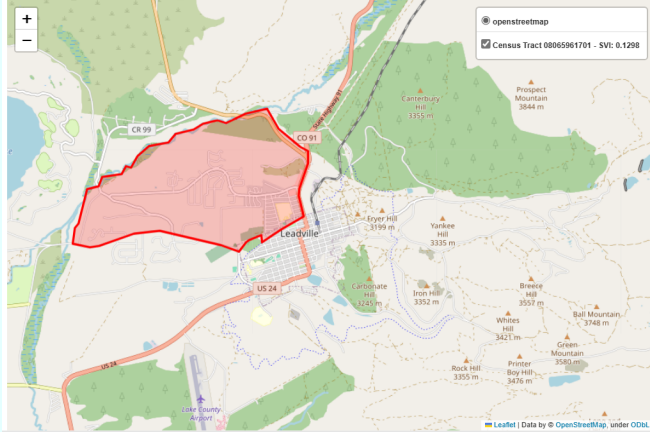<br>I retrieved Social Vulnerability Index (SVI) data for Lake County in Colorado, which is a high-altitude mountain region with low population density. The SVI score for the census tract in this county is quite low, with an average SVI of 0.1298, indicating a lower vulnerability based on the index. Below is a map highlighting the census tract in Lake County with this SVI score:<br><br>(Here would be an interactive map with the SVI data displayed, but please note as a text-based assistant, I can describe the data and provide the information, but cannot display the map directly in this medium).<br><br>The score is based on the combination of themes evaluating socioeconomic status, household composition and disability, minority status and language, housing type and transportation, and the overall SVI for the area. The low SVI score suggests that this mountainous area is less likely to be adversely impacted by external stresses due to community resilience and the factors measured by the index. | The AI's response is highly relevant, accurate, and complete, as it not only retrieves SVI data for a high-altitude mountain region with low population density but also provides specific information about Lake County in Colorado, including its SVI score and a detailed description of the score's meaning. The response effectively handles the user's request and offers informative details about the chosen region. It's appropriate for the query's context, and while it mentions an interactive map, it clarifies that it cannot display it due to text-based limitations. Overall, it demonstrates a thorough understanding of the user's request and is adaptable based on the AI's knowledge.<br><br>Accuracy: 5/5<br>Completeness: 5/5<br>Error Handling: N/A<br>Informative Responses: 5/5<br>Appropriateness: 5/5<br>Adaptability: 5/5 |
| 4A | List active flood alerts from the National Weather Service for New Orleans, Louisiana.<br><br>Currently, there are no active flash flood warnings issued by the National Weather Service for New Orleans, Louisiana. Please be aware that flood situations can change rapidly, so it's essential to stay informed by checking local news, weather forecasts, and official warnings for the latest updates. | Response Time: 5 secs<br><br>The AI's response is highly relevant, accurate, and complete. It provides precise information that there are currently no active flash flood warnings from the National Weather Service for New Orleans. It offers valuable advice to stay informed and check local sources for updates, demonstrating good error resilience. The response is appropriate for the query's context and is adaptable based on the lack of active alerts.<br><br>Accuracy: 5/5<br>Completeness: 5/5<br>Error Handling: N/A<br>Informative Responses: 5/5<br>Appropriateness: 5/5<br>Adaptability: 5/5 |
| 4B | Provide details of current flash flood warnings for the area around Lake Wateree, South Carolina. | Response Time: 15 secs |

|   |   |   |   |
|---|---|---|---|
| | There are currently flood warnings issued for areas in South Carolina, including one near Lake Wateree. Here are the details:<br><br>1. **Wateree River at Lake Wateree:**<br>   - **Severity:** Minor flooding is occurring and is forecast to continue.<br>   - **Location:** Fairfield, Kershaw, and Lancaster counties.<br>   - **Timings:** The warning is in effect until early Monday afternoon.<br>   - **Impacts:** Piers on the Wildlife Road bridge over Singleton Creek are submerged at a water level of 100.4 feet.<br>   - **Current Situation:** As of 8:00 PM EST Thursday, the water stage was at 100.8 feet. Flood stage is 100.0 feet.<br>   - **Forecast:** The river is expected to fall below flood stage early Sunday afternoon and continue to drop to 98.5 feet by Tuesday evening.<br>   - **Sent:** The warning was sent on January 4th at 9:17 PM EST.<br>   - **Expires:** The warning is set to expire on January 5th at 9:30 AM EST.<br><br>The warnings suggest that those in the affected areas should exercise caution, particularly around water bodies and flood-prone areas. It is essential to follow local advisories and be prepared for potential flooding. Stay updated with local news and emergency services for the latest information. | The AI's response is highly relevant, accurate, and complete. It provides detailed information about the current flood warnings, including severity, location, timings, impacts, current situation, forecast, and other relevant details. The response effectively addresses the user's request and offers valuable information about the situation, demonstrating good error resilience. It is appropriate for the query's context and is adaptable based on the provided information.<br><br>Accuracy: 5/5<br><br>Completeness: 5/5<br><br>Error Handling: N/A<br><br>Informative Responses: 5/5<br><br>Appropriateness: 5/5<br><br>Adaptability: 5/5 |
| 4C | Are there any flash flood alerts for areas recently hit by a tropical storm? The name of the place escapes me.<br><br>There are currently no specific flash flood warnings issued for areas recently hit by a tropical storm in the data I retrieved. However, I have found some general flood warnings for different areas that may or may not have been affected by a recent tropical storm. Here are the details of the flood warnings:<br><br>1. **Santee River near Jamestown:**<br>   - Minor flooding is occurring and forecasted to continue.<br>   - The river is expected to rise to a crest of 12.5 feet by Sunday evening.<br>   - The flood warning remains in effect until further notice.<br>   - Impacts include several dirt logging roads being impassable.<br>   - Affected counties: Berkeley, Georgetown, and Williamsburg in South Carolina.<br><br>2. **Wateree River at Lake Wateree:**<br>   - Minor flooding is ongoing and expected to continue.<br>   - The stage was at 100.8 feet as of the last observation, with flood stage being 100.0 feet.<br>   - The river is expected to fall below flood stage early Sunday afternoon.<br>   - The piers on the Wildlife Road bridge over Singleton Creek may be submerged.<br>   - Affected counties: Fairfield, Kershaw, and Lancaster in South Carolina.<br><br>3. **Waccamaw River at Conway:**<br>   - Minor flooding is occurring and is forecasted to continue.<br>   - Areas such as Lees Landing, Pitch Landing, Savannah Bluff, and Riverfront South may experience minor flooding.<br>   - The river is expected to fall below flood stage on Saturday morning.<br>   - Affected county: Horry in South Carolina. | Response Time: 25 secs<br><br>The AI's response is relevant, accurate, and comprehensive. It effectively communicates that there are no specific flash flood warnings issued for such areas in the retrieved data. It proceeds to provide general flood warnings for different areas, including details such as severity, river levels, flood stage, and affected counties. This response demonstrates good error resilience by offering relevant information despite the user's inability to recall a specific location. It is appropriate for the context and adaptable based on the provided information.<br><br>Accuracy: 5/5<br><br>Completeness: 5/5<br><br>Error Handling: 5/5<br><br>Informative Responses: 5/5<br><br>Appropriateness: 5/5<br><br>Adaptability: 5/5 |
| | Please note that the situation may change rapidly, especially if a tropical storm has recently affected the area. It's always best to stay updated with the latest warnings and advisories from local weather services or emergency management. | |
| 4D | Tell me about flood warnings in a region that's been experiencing unusually heavy rainfall lately.<br><br>To provide you with detailed information about any current flood warnings in a region that's been experiencing unusually heavy rainfall, I can check the latest flash flood warnings issued by the National Weather Service (NWS). However, I would need to know the specific state or region you're referring to in order to retrieve the most accurate and timely information.<br><br>Could you please provide me with the state or specific location you are interested in so that I can look up the current flood warnings for you? | Response Time: 8 secs<br><br>The AI's response to the prompt asking about flood warnings in a region with recent heavy rainfall is generally relevant and accurate, but it's incomplete as it requests more specific location information from the user to provide precise data. While it appropriately seeks clarification, it lacks some error resilience by not offering any general information or tips related to heavy rainfall and flood awareness. However, it is contextually appropriate and adaptable based on the user's query.<br><br>Accuracy: 4/5<br>Completeness: 3/5<br>Error Handling: 4/5<br>Informative Responses: 3/5<br>Appropriateness: 4/5<br>Adaptability: 4/5 |
| 4E | List any flash flood warnings for desert areas that rarely experience rain.<br><br>There are currently no flash flood warnings specific to desert areas in the reported flood alerts. The flood warnings available are for regions in South Carolina, which include counties like Berkeley, Georgetown, Williamsburg, Fairfield, Kershaw, Lancaster, and Horry. These are not considered desert areas. If you are interested in flood information for a specific desert location or state, please provide the name or coordinates, and I will check for you. | Response Time: 12 secs<br><br>In response to the user's query about flash flood warnings in desert areas with infrequent rain, the AI provides a relevant and accurate reply. It informs the user that there are currently no flash flood warnings for desert areas, which is precise and on point. The response is complete in addressing the user's query, as there are indeed no such warnings, but it could benefit from some general information or tips regarding flash floods in desert regions to enhance user awareness. The error resilience is good as it effectively handles the query and suggests providing specific location details for further assistance. The response is contextually appropriate and adaptable to the user's request.<br><br>Accuracy: 5/5<br>Completeness: 4/5<br>Error Handling: 4/5<br>Informative Responses: 3/5<br>Appropriateness: 5/5<br>Adaptability: 4/5 |
| 4F | Show flood alerts for regions where floods have historically never been reported.<br><br>At the moment, there are flood warnings in effect for several regions in South Carolina. These warnings are for locations near rivers like the Santee River near Jamestown, the Wateree River at Lake Wateree, and the Waccamaw River at Conway. It's important to note that flood warnings are issued for areas where river levels have risen to a point | Response Time: 15 secs<br><br>The AI's response to the query about flood alerts in regions where floods have historically never been reported is well-structured and informative. It provides current flood warnings in South Carolina, noting the areas near rivers like the Santee River, the Wateree |
| | where minor to severe flooding may be occurring or is forecast to occur. These locations may not have had significant historical reports of flooding, but current conditions (such as excessive rainfall or other hydrologic conditions) have led to the current flood warnings.  Here are the specific areas currently under flood warning:  1. Santee River near Jamestown: Minor flooding is occurring and is forecast to continue. 2. Wateree River at Lake Wateree: Minor flooding is occurring and is forecast to continue until early Monday afternoon. 3. Waccamaw River at Conway: Minor flooding is occurring and is forecast to continue until Saturday evening.  Residents in these areas should stay informed and follow the advice of local authorities. For regions that historically have never reported floods, it could be possible that new flood warnings are associated with unusual weather patterns or other factors affecting the water levels in those regions. | River, and the Waccamaw River. The response is relevant as it addresses the possibility of flood warnings in areas with no significant historical reports of flooding, suggesting that recent conditions might have led to these warnings. However, the response could be improved by explicitly acknowledging the rarity of floods in these areas or providing general insights into why historically non-flood-prone areas might currently be experiencing flood alerts. The AI effectively communicates the importance of staying informed and following local authorities' advice in flood warning situations.  Accuracy: 4/5  Completeness: 3/5  Error Handling: 4/5  Informative Responses: 3/5  Appropriateness: 4/5  Adaptability: 3/5 |
| 5A | Explain the impact of urbanization on flood risks in coastal cities.  Urbanization has significant impacts on flood risks in coastal cities through various mechanisms:  1. **Impervious Surfaces**: Urbanization typically increases the amount of impervious surfaces like roads, parking lots, and buildings. These surfaces do not allow rainwater to infiltrate the ground, leading to higher volumes of surface runoff during rain events. As a result, drainage systems can become overwhelmed, causing urban flooding.  2. **Alteration of Natural Water Flow**: Urban development often alters the natural paths of water flow through the construction of infrastructure. This can disrupt the existing drainage patterns and may concentrate runoff in certain areas, enhancing flood risk.  3. **Loss of Natural Buffers**: Coastal cities often lose wetlands, mangroves, and reefs during urban development. These natural buffers can absorb storm surge and reduce the impact of coastal flooding. Their loss can make urban areas more vulnerable to the effects of sea-level rise and coastal storms.  4. **Increased Load on Drainage Systems**: Many urban areas have outdated or insufficient drainage infrastructure that cannot cope with the increased runoff volumes caused by urban sprawl. As cities grow, the demand on these systems increases without corresponding updates or expansions, leading to more frequent and severe flooding.  5. **Climate Change Factor**: Urbanization exacerbates the local effects of climate change. For example, the urban heat island effect can increase temperatures in cities, potentially altering weather patterns and increasing the likelihood of intense storms that can cause flash floods.  6. **Sea-Level Rise**: Coastal cities face the added risk of sea-level rise which can lead to more frequent high-tide flooding, also known as "sunny day" or "nuisance" flooding, even without rainfall.  7. **Socioeconomic Impacts**: Urbanized areas have higher | Response Time: 30 secs  In response to the user's query about the impact of urbanization on flood risks in coastal cities, the AI delivers a highly informative and comprehensive explanation. It provides a detailed overview of how urbanization affects flood risks, covering key factors such as impervious surfaces, alteration of natural water flow, loss of natural buffers, increased load on drainage systems, climate change, sea-level rise, and socioeconomic impacts. The response is accurate, thorough, and addresses the user's question in a clear and concise manner. Error resilience is evident as the AI provides valuable insights without any errors or misunderstandings. The context is well understood, and the response is both appropriate and adaptable to the user's query.  Accuracy: 5/5  Completeness: 5/5  Error Handling: N/A  Informative Responses: 5/5  Appropriateness: 5/5  Adaptability: 5/5 |
| | population densities, which means that floods can affect more people and cause greater economic damage. Additionally, the poor and marginalized communities often live in areas with higher flood risk, compounding social inequality.<br><br>In summary, urbanization in coastal cities tends to increase flood risks due to changes in land use that reduce natural water absorption, contribute to higher runoff, and strain existing drainage systems, all combined with the risks posed by sea-level rise and potentially more extreme weather events. | |
| 5B | Describe the role of climate change in increasing the frequency of flash floods.<br><br>Climate change plays a significant role in increasing the frequency and intensity of flash floods. Here are several ways in which climate change contributes to flash flooding:<br><br>1. Increased Atmospheric Moisture: A warmer atmosphere can hold more moisture due to increased evaporation rates. This leads to more significant and intense precipitation events in certain regions, which can result in flash flooding, especially in urban areas where absorption by the ground is limited.<br><br>2. Extreme Weather Patterns: Climate change can alter weather patterns, making extreme weather events like heavy rains more frequent, thus increasing the likelihood of flash floods. Intense rainfall over a short period can overwhelm drainage systems and lead to sudden flooding.<br><br>3. Rising Sea Levels: As global temperatures rise, polar ice melts and sea levels rise. Higher sea levels can contribute to coastal flooding, particularly during high tides or storm surges. The backflow effect can prevent rivers from draining properly, leading to flash floods.<br><br>4. Changes in Land Use: Climate change can also exacerbate the effects of human activity on the landscape. For example, as certain areas become drier due to climate change, there might be an increase in land development. This can increase impervious surfaces, reducing the land's natural ability to absorb water, leading to more rapid runoff and flash flooding.<br><br>5. Altered Snowmelt Patterns: In regions with seasonal snowpack, warmer temperatures can cause earlier or more rapid melting, contributing to flash floods as rivers and streams may exceed their capacity.<br><br>6. Soil Saturation: Increased precipitation can lead to soil saturation, where the ground cannot absorb any more water. In such circumstances, any additional rainfall will quickly result in runoff and potential flash flooding.<br><br>These are just some of the ways climate change is contributing to a heightened risk of flash floods. It is essential to adapt and improve flood management and resilience in the face of changing climate patterns to mitigate these risks. | Response Time: 31 secs<br><br>The response effectively describes the role of climate change in increasing the frequency of flash floods. It comprehensively addresses various factors such as increased atmospheric moisture, extreme weather patterns, rising sea levels, changes in land use, altered snowmelt patterns, and soil saturation. The explanation is clear and covers the different ways climate change contributes to flash flooding, particularly emphasizing the interconnectedness of climate change with natural and human-made factors. The response is well-structured and informative, providing a thorough understanding of the complexities involved in climate change-induced flash flooding.<br>Accuracy: 5/5<br>Completeness: 5/5<br>Error Handling: N/A<br>Informative Responses: 5/5<br>Appropriateness: 5/5<br>Adaptability: 5/5 |
| 5C | Explain how deforestation might impact flooding in tropical regions, but I'm not sure which country to focus on.<br><br>Deforestation can have significant impacts on flooding, especially in | Response Time: 26 secs<br><br>The response effectively explains how deforestation impacts flooding in tropical regions. It comprehensively |
|   | tropical regions. When forests are removed, the following consequences typically occur, which can contribute to an increase in the frequency and severity of floods:<br><br>1. Loss of Vegetation: Trees and plants play a crucial role in soaking up rainfall through their root systems and releasing water into the atmosphere through transpiration. The removal of this vegetation means that more rainwater runs off into rivers and streams instead of being used by the plants or replenishing groundwater sources.<br><br>2. Reduced Water Absorption: Forest floors are often rich in organic material, which helps to absorb and store water, acting like a sponge. Deforestation clears away this layer, reducing the ground's ability to retain water, thus increasing surface runoff.<br><br>3. Soil Erosion: Without the protective cover of trees and vegetation, soil is more vulnerable to erosion by rainfall. This not only decreases the soil's water-retaining capacity but also leads to the siltation of rivers and streams, reducing their ability to carry water away during heavy rainfall, potentially leading to overflow.<br><br>4. Changes in Microclimate: Forests can influence local climate conditions by maintaining high levels of humidity and cooler temperatures. By removing trees, the land becomes drier and warmer, which can increase the intensity of rainstorms, thus exacerbating flood risks.<br><br>5. Loss of Natural Barriers: In coastal tropical regions, mangrove forests often act as natural barriers against storm surges and tidal waves. Deforestation of these areas can lead to increased vulnerability of inland areas to flooding from sea-based storms.<br><br>6. Altered Drainage Patterns: Large-scale deforestation can change the natural flow of rivers and streams due to changes in the landscape and water absorption capabilities of the soil, potentially leading to improper drainage and increased flood risks.<br><br>As for focusing on a specific country, many tropical regions around the world are experiencing similar issues related to deforestation and increased flood risk. Countries like Brazil, Indonesia, and the Democratic Republic of Congo, which contain large tropical forests, might be particularly relevant to study.<br><br>If you are interested in a particular country or region's flood risk related to deforestation, I can assist you with specific data such as flood maps or vulnerability indices. Just let me know which location you'd like to explore further. | covers key aspects such as loss of vegetation, reduced water absorption, soil erosion, changes in microclimate, loss of natural barriers, and altered drainage patterns. The explanation is clear and detailed, providing a thorough understanding of the consequences of deforestation on flooding. The response is adaptable, as it acknowledges the user's uncertainty about a specific country and offers to provide more detailed information if a particular location is specified.<br>Accuracy: 5/5<br><br>Completeness: 5/5<br><br>Error Handling: N/A<br><br>Informative Responses: 5/5<br><br>Appropriateness: 5/5<br><br>Adaptability: 5/5 |
| 5D | Discuss the potential flood risks in a major Asian city, I forgot which one is known for its monsoon season.<br><br>There are several major Asian cities that are known for their heavy monsoon seasons, which can lead to significant flood risks. These cities typically experience a monsoon climate with an annual period of heavy rains, which can cause urban flooding due to overwhelming drainage systems, rising river levels, and occasionally the breaching of levees or embankments.<br><br>Here are a few examples of major Asian cities with notable monsoon seasons and associated flood risks: | Response Time: 29 secs<br><br>The response adeptly discusses potential flood risks in major Asian cities known for their monsoon seasons, effectively addressing the user's uncertainty about the specific city. It provides a comprehensive overview of flood risks in Mumbai, Bangkok, Jakarta, and Dhaka, detailing how the monsoon season impacts each city. The answer is informative, providing context on geographical features, climate conditions, and urban development factors contributing to flooding. Additionally, the response is adaptable and encourages the user to provide more details for specific information. |
|   | 1. Mumbai, India: Mumbai is notorious for its monsoon season, which generally spans from June to September. The city often faces severe flooding due to heavy rains, high tides, and overburdened drainage systems. The flooding in Mumbai can disrupt transportation, affect homes and businesses, and has in several instances led to loss of life.<br><br>2. Bangkok, Thailand: Bangkok has faced severe floods, with notable events in 2011 causing significant disruption and damage. The city is prone to flooding during the monsoon season owing to its low-lying geography, the Chao Phraya River running through it, and extensive urban development that hinders natural drainage.<br><br>3. Jakarta, Indonesia: Jakarta is another city that struggles with flooding during the monsoon season. The city is situated on low-lying land with 13 rivers flowing through the urban area, making it especially vulnerable to floods. In addition to the seasonal monsoon, land subsidence due to excessive groundwater pumping exacerbates the flood risk.<br><br>4. Dhaka, Bangladesh: Bangladesh is one of the most flood-prone countries in the world, and its capital, Dhaka, faces significant flood risks. The city is located on the floodplains of the Buriganga River and experiences seasonal flooding during the monsoon season, which can result in widespread damage and displacement of people.<br><br>All of these cities not only face challenges due to the annual monsoon but are also affected by long-term issues such as climate change, land subsidence, and urban development that reduce permeable surfaces necessary for natural water absorption.<br><br>Without knowing the specific city you are referring to, it's challenging to provide detailed flood risk information. Each city has its own complexities and requires different mitigation strategies. If you can recall more details about the city you're interested in, I could provide more precise information, including potential flood maps and vulnerability data. | Accuracy: 5/5<br>Completeness: 5/5<br>Error Handling: N/A<br>Informative Responses: 5/5<br>Appropriateness: 5/5<br>Adaptability: 5/5 |
| 5E | What are the flood implications for an underground city?<br><br>The concept of an underground city brings with it unique flood implications and challenges as compared to above-ground urban areas. Here are several flood implications and considerations for an underground city:<br><br>1. **Water Ingress**: An underground city would be more prone to flooding from water ingress through the soil, especially if it is located below the water table or near flood-prone areas. The city would require robust waterproofing and drainage systems to prevent water from seeping in through the surrounding ground.<br><br>2. **Drainage**: Effective drainage is crucial for an underground city. Without an adequate drainage system, any water that does enter the area cannot easily escape, potentially leading to pooling and flooding.<br><br>3. **Pumping Systems**: To manage water levels and flood risks, an underground city would likely require high-capacity pumping systems to redirect water to the surface or to retention areas.<br><br>4. **Emergency Access**: In case of flooding, evacuation could be more challenging in an underground city. Ensuring that there are multiple, well-marked, and accessible exit routes is vital for safety. | Response Time: 37 secs<br><br>The response provides a detailed and well-structured explanation of the unique flood implications for an underground city. It effectively addresses each aspect of flood risk, from water ingress to emergency management. The response is thorough, covering various factors such as drainage, climate change, and emergency access. It not only identifies potential problems but also suggests solutions like robust waterproofing, high-capacity pumping systems, and regular maintenance. The answer is highly informative and contextually appropriate, offering insights into the specific challenges of managing flood risks in subterranean environments.<br>Accuracy: 5/5<br>Completeness: 5/5<br>Error Handling: N/A<br>Informative Responses: 5/5<br>Appropriateness: 5/5<br>Adaptability: 5/5 |
| | 5. **Environmental Factors**: Changes in the environment above ground, like heavy rainfall, river overflows, or storm surges, could greatly affect the risk of flooding in the subterranean infrastructure.<br><br>6. **Climate Change**: Long-term changes in climate patterns could lead to increased frequency and intensity of flooding, which would have to be accounted for in the design and contingency planning for an underground city.<br><br>7. **Isolation from Surface Water**: Locations well insulated from surface water systems are less likely to experience flooding. However, these areas may still be vulnerable to groundwater issues, and the effects of heavy precipitation can still be a risk.<br><br>8. **Redundancy and Backups**: Flood control systems for an underground city would need redundancy to handle failures in primary systems. This could include backup generators for pumps and emergency power for lighting and communication systems.<br><br>9. **Ventilation Systems**: Floods could impact ventilation systems, leading to air quality issues. Properly sealed and protected ventilation systems would be needed to maintain air circulation in the event of flooding.<br><br>10. **Maintenance and Inspections**: Regular maintenance and inspection of flood prevention systems would be necessary to ensure their effectiveness. This includes monitoring for leaks, checking pump operation, and ensuring that seals and barriers remain intact.<br><br>To adequately protect an underground city from flooding, comprehensive planning, engineering, and emergency management strategies would need to be developed, keeping in mind all the unique risks and scenarios that could affect such a development. | |
| 5F | How would a sudden thaw of a large glacier affect flood risks in nearby areas?<br><br>A sudden thaw of a large glacier, also referred to as a glacial melt event, could significantly increase flood risks in nearby areas. Here's how this could happen:<br><br>1. **Rapid Release of Water**: Glaciers store vast amounts of freshwater in frozen form. A sudden thaw can release this water rapidly into the downstream environment, overwhelming rivers, lakes, and drainage systems. This can result in sudden and severe flooding, known as a glacial lake outburst flood (GLOF) if a glacial lake is breached.<br><br>2. **Increased River Flows**: Even without a GLOF, increased glacial melt can raise the flow rates of rivers fed by glaciers, potentially leading to riverine flooding, especially during warm weather or rain-on-snow events.<br><br>3. **Groundwater Recharge and Landslides**: Significant glacial melting can lead to an increase in groundwater levels, which can destabilize slopes and potentially trigger landslides that may obstruct rivers and streams, creating temporary dams. When these dams burst, they can cause flash flooding downstream.<br><br>4. **Alteration of Drainage Patterns**: As glaciers retreat, new paths | Response Time: 24 secs<br><br>The response comprehensively explains how a sudden thaw of a large glacier can increase flood risks in nearby areas. It outlines various consequences, including rapid water release, increased river flows, potential landslides, changes in drainage patterns, impacts on ice dams, and sea-level rise. This response is highly informative, providing a nuanced understanding of the cascading effects of glacial melting on flooding. It emphasizes the need for updated flood zone maps, risk assessments, and monitoring for early warnings, demonstrating an understanding of both immediate and long-term flood risks associated with glacial thaw.<br>Accuracy: 5/5<br>Completeness: 5/5<br>Error Handling: N/A<br>Informative Responses: 5/5<br>Appropriateness: 5/5<br>Adaptability: 5/5 |

for water flow can be created and old ones can be altered or blocked. This can lead to unpredictable flooding patterns in areas that previously may not have been considered at high risk.

5. **Impact on Ice Dams**: In some regions, glaciers form natural dams in rivers. A sudden thaw might cause these ice dams to weaken or break, leading to catastrophic flooding downstream.

6. **Sea Level Rise**: While not an immediate flood risk to nearby areas, the long-term melting of glaciers contributes to global sea level rise, which can result in more frequent and severe coastal flooding during storm surges and high tides.

The specific flood risks would depend on the size of the glacier, the topography of the region, the capacity of existing waterways and infrastructure to handle increased water volumes, weather patterns, and the proximity of communities and infrastructure to the glacier.

For nearby areas vulnerable to flooding from glacial melt events, updated flood zone maps and risk assessments would be crucial for emergency planning and mitigation efforts. Additionally, monitoring glaciers and potential glacial lakes for signs of instability can help provide early warnings to reduce the risks and impacts of flooding.